%% file: main.tex
\documentclass[letterpaper,journal]{IEEEtran}
\usepackage{amsmath,amsfonts}
\usepackage{algorithm}
\usepackage[noend]{algpseudocode}

\usepackage{array}
\usepackage[caption=false,font=normalsize,labelfont=sf,textfont=sf]{subfig}
\usepackage{textcomp}
\usepackage{stfloats}
\usepackage{url}
\usepackage{verbatim}
\usepackage{graphicx}
\usepackage{cite}

\usepackage{pgfplots}
\usepackage{cite}
\usepackage{graphicx}
\usepackage{amsmath,amssymb,amsfonts}
\usepackage{algorithm}
\usepackage{algpseudocode}
\usepackage{booktabs}
\usepackage{textcomp}
\usepackage{xcolor}
\def\BibTeX{{\rm B\kern-.05em{\sc i\kern-.025em b}\kern-.08em
    T\kern-.1667em\lower.7ex\hbox{E}\kern-.125emX}}
\usepackage{multirow}
\usepackage{algorithmicx}
\usepackage{algpseudocode}
\usepackage{tabularx}
\graphicspath{{Fig/}}
\usepackage{amsthm}
\usepackage{pdfpages}
\usepackage{subfiles}
\usepackage{multicol}

\usepackage{tikz}
\usetikzlibrary{shapes.geometric, arrows.meta, positioning}
\usetikzlibrary{shapes.geometric, arrows, positioning}
\tikzstyle{line} = [draw, -latex']
\tikzset{
  block/.style={rectangle, draw, fill=white, 
                text width=4.5cm, align=center, 
                rounded corners, minimum height=1cm},
  line/.style={draw, -Latex}
}
\theoremstyle{thmstyleone}%
\theoremstyle{thmstyletwo}%
\theoremstyle{thmstylethree}%

\author{
    \IEEEauthorblockN{Mustafa Mohammadi Gharasuie\IEEEauthorrefmark{1}, Luis Rueda\IEEEauthorrefmark{2}}
    \\\IEEEauthorrefmark{1}\IEEEauthorrefmark{2} School of Computer Science, University of Windsor, Windsor, ON, Canada
    \\\IEEEauthorblockA{\IEEEauthorrefmark{1}Mohamm6m@uwindsor.ca
    \\\IEEEauthorrefmark{2}lrueda@uwindsor.ca}
}

\begin{document}

\title{Accelerating Image Classification with Graph Convolutional Neural Networks using Voronoi Diagrams}

\markboth{Journal of IEEE Transactions on Image Processing}
{Shell \MakeLowercase{\textit{et al.}}: A Sample Article Using IEEEtran.cls for IEEE Journals}


\maketitle

\begin{abstract}
Recent advances in image classification have been significantly propelled by the integration of Graph Convolutional Networks (GCNs), offering a novel paradigm for handling complex data structures. This study introduces an innovative framework that employs GCNs in conjunction with Voronoi diagrams to peform image classification, leveraging their exceptional capability to model relational data. Unlike conventional convolutional neural networks, our approach utilizes a graph-based representation of images, where pixels or regions are treated as vertices of a graph, which are then simplified in the form of the corresponding Delaunay triangulations. 
Our model yields significant improvement in pre-processing time and classification accuracy on several benchmark datasets, surpassing existing state-of-the-art models, especially in scenarios that involve complex scenes and fine-grained categories. The experimental results, validated via cross-validation, underscore the potential of integrating GCNs with Voronoi diagrams in advancing image classification tasks. This research contributes to the field by introducing a novel approach to image classification, while opening new avenues for developing graph-based learning paradigms in other domains of computer vision and non-structured data. In particular, we have proposed a new version of the GCN in this paper, namely normalized Voronoi Graph Convolution Network (NVGCN), which is faster than the regular GCN.
\end{abstract}


\section{Introduction}
The domain of image classification has witnessed a paradigm shift with the advent of deep learning techniques, particularly Graph Convolutional Neural Networks (GCNs), which have revolutionized how we approach complex tasks on image data. GCNs, by their nature, are adept at handling data represented in a graph format, making them an ideal choice for tasks where relational context and structural information are pivotal. This introduction outlines recent advancements in image classification using GCNs, focusing on the use of superpixels and wavelet techniques.

GCNs have emerged as a powerful paradigm in image classification, primarily due to their ability to capture and process the non-Euclidean structure of the data. Zhou, et al. provided a comprehensive overview of the applications of GCNs in various fields, emphasizing their efficacy in image classification tasks \cite{zhou2020graph}. They highlighted how GCNs, unlike traditional CNNs, can capture long-range dependencies and complex relational patterns in images.
Wavelet techniques, on the other hand, have revolutionized the field of image processing by providing a multi-resolution analysis framework critical for various applications such as image compression, feature extraction, and noise reduction.

In this context, Wavelet transforms have been utilized in conjunction with GCNs for feature extraction in image classification, denoising, compression and other tasks. Wavelets provide a multi-resolution image analysis, capturing spatial and frequency domain information \cite{mallat1989theory}. 

 This paper introduces an innovative framework that employs GCNs in conjunction with Voronoi Diagrams for image classification, leveraging their exceptional capability to model relational data. Unlike conventional convolutional neural networks, our approach utilizes the graph-based representation of images in the form of superpixels, and their relational contexts are captured as edges. 
 Incorporating Voronoi diagrams aids in partitioning the image into distinct, yet interrelated regions, facilitating an enriched, context-aware graph construction. The method allows for a more nuanced understanding of spatial and contextual relationships within the image data. In contrast, we propose a unique graph construction technique that effectively captures both local and global features of the underlying images, enhancing the model's ability to discern intricate patterns and subtle distinctions. The main contributions of this paper are the following: 
 (i) converting an image into graphs via superpixels without needing to pre-process the image; 
 (ii) introducing a new form of image representation via a Voronoi diagrams in which the superpixels are the regions; 
 (iii) developing a model for dual graph representation of the images via a Delaunay triangulations; 
 (iv) introducing a new version of Graph Convolution Networks, NVGCN which run much faster than the conventional GCN in terms of time complexity and the number of multiplications in each layer.
 %

\section{Related Work}
Simple Linear Iterative Clustering (SLIC) is a method proposed in \cite{8-SLIC-Achanta2012SLIC}, which aims at creating superpixels by grouping pixels together.
\\
The task is based on two main criteria: their color similarity and spatial proximity. The algorithm's core lies in minimizing a distance measure, $D$, which is a combination of color distance $d_{lab}$ in the CIELAB color space and spatial distance $d_{xy}$ on the image coordinate space. The composite distance measure used by SLIC to form superpixels is given by:
\begin{equation}
    D = \sqrt{(d_{lab})^2 + \left(\frac{d_{xy}}{S}\right)^2 \cdot m^2}     \, ,
\label{eq:SLIC_formula}
\end{equation}
Here, $S$ is the spacing between superpixel centers, and $m$ is a compactness parameter controlling the trade-off between color similarity and spatial proximity. The algorithm iteratively refines superpixel boundaries by recalculating the distance $D$ and adjusting the pixel assignments to the nearest cluster center, with the process repeating until convergence.

Another approach is Manifold SLIC, proposed in \cite{8-SLIC-Achanta2012SLIC}, which is an adaptation of the Simple Linear Iterative Clustering (SLIC) algorithm and is designed to work with data that lies on a manifold rather than in a Euclidean space. In image processing, a manifold can be considered a curved surface in a high-dimensional space, where the manifold's intrinsic geometry captures the underlying structure of the image data more effectively than a flat Euclidean representation.

A general representation of the manifold distance \( D_m \) in Manifold SLIC can be expressed as follows:

\begin{equation}
    D_m = \sqrt{(d_{lab})^2 + \left(\frac{d_{geo}}{S}\right)^2 \cdot m^2} \, ,    
\label{eq:ManifoldSLIC_formula}
\end{equation}

In this formula, \( d_{lab} \) represents the color distance in the CIELAB space, while \( d_{geo} \) is the geodesic distance on the manifold, which measures the shortest path between two points along the manifold's surface. \( S \) is the grid interval for superpixels, and \( m \) is a parameter that balances color similarity against spatial proximity, similar to the compactness parameter in the original SLIC.


Another advancement in superpixel segmentation methods is the \emph{Superpixels None-Iterative Clustering} (SNIC) algorithm \cite{15-SNIC-achanta2017superpixels}. SNIC builds upon the foundation laid by SLIC, though it introduces several improvements to enhance performance and computational efficiency. Unlike SLIC, which initializes cluster centers and iteratively refines them, SNIC starts by treating each pixel as a separate cluster and progressively merges them based on color and spatial proximity. This approach ensures that the generated superpixels adhere more closely to object boundaries while maintaining its computational simplicity.

Furthermore, SNIC addresses some of SLIC's limitations regarding connectivity and execution speed. By employing a priority queue to manage the growth of the corresponding superpixels, SNIC guarantees connected regions and reduces computational complexity. Experimental results have demonstrated that SNIC achieves comparable or superior segmentation quality compared to SLIC, with faster execution times~\cite{15-SNIC-achanta2017superpixels}. Therefore, SNIC represents a significant step forward in efficient and accurate superpixel generation, making it a valuable tool for various computer vision applications.

TurboPixels is a technique introduced in \cite{9-turbopixel}. It consists of an algorithm for generating superpixels, which are small, coherent regions in an image. The technique is designed to produce superpixels that adhere closely to the boundaries in the image while maintaining a uniform size and shape. The main idea TurboPixels is to grow contours that start from an initial seed point and grow iteratively to cover the entire image. In the growth process of TurboPixels, a key element is the geodesic distance, which ensures that the superpixel boundaries align with image contours. The main formula that governs the evolution of the contour is typically a partial differential equation (PDE) based on the level set method. This PDE is formulated to minimize an energy functional \( E \) that combines the image's intensity gradient and the contour's curvature. While the specific formula can be quite complex, involving calculus of variations, can be expressed as follows:

\begin{equation}
    E(C) = \int_{C} g(I(x, y)) \cdot \lVert \nabla C(x, y) \rVert + \alpha \cdot \kappa(C(x, y)) \, ds
\label{eq:TurboPixel_formula}
\end{equation}

In this formula, \( C \) represents the contour, \( g(I(x, y)) \) is a stopping function derived from the image intensity \( I \) that slows the contour evolution near strong edges, \( \nabla C \) is the gradient of the contour, \( \kappa(C(x, y)) \) is the curvature of the contour, \( \alpha \) is a weighting factor that controls the tradeoff between adherence to image boundaries and contour smoothness, and the integration is carried out over the contour length.


Another approach is Linear Spectral Clustering (LSC), proposed in \cite{11-LinearSpectralClustering7814265}, which is an advanced technique designed for image segmentation that leverages the power of spectral clustering methods. It efficiently segments an image by mapping pixels into a lower-dimensional space where clusters are more easily identifiable. 

The core operation in LSC involves constructing a similarity matrix \(W\) and then solving for the eigenvectors of the Laplacian matrix \(L = D - W\), where \(D\) is the diagonal matrix with degrees of the vertices. The main formula used to identify clusters within the transformed space is derived from the eigenvalue problem:
\begin{equation}   
    L_\mathbf{u} = \lambda D_\mathbf{u} \, .
\label{eq:LSC_formula}
\end{equation}
Here, \(\mathbf{u}\) represents the eigenvector corresponding to the eigenvalue \(\lambda\), and solving this problem yields a set of eigenvectors that are then used for clustering. The smallest non-zero eigenvalues and their corresponding eigenvectors capture the most significant relationships between pixels, which are indicative of the underlying segments within the image.

\begin{figure}
    \centering
    \includegraphics[width=0.5\textwidth]{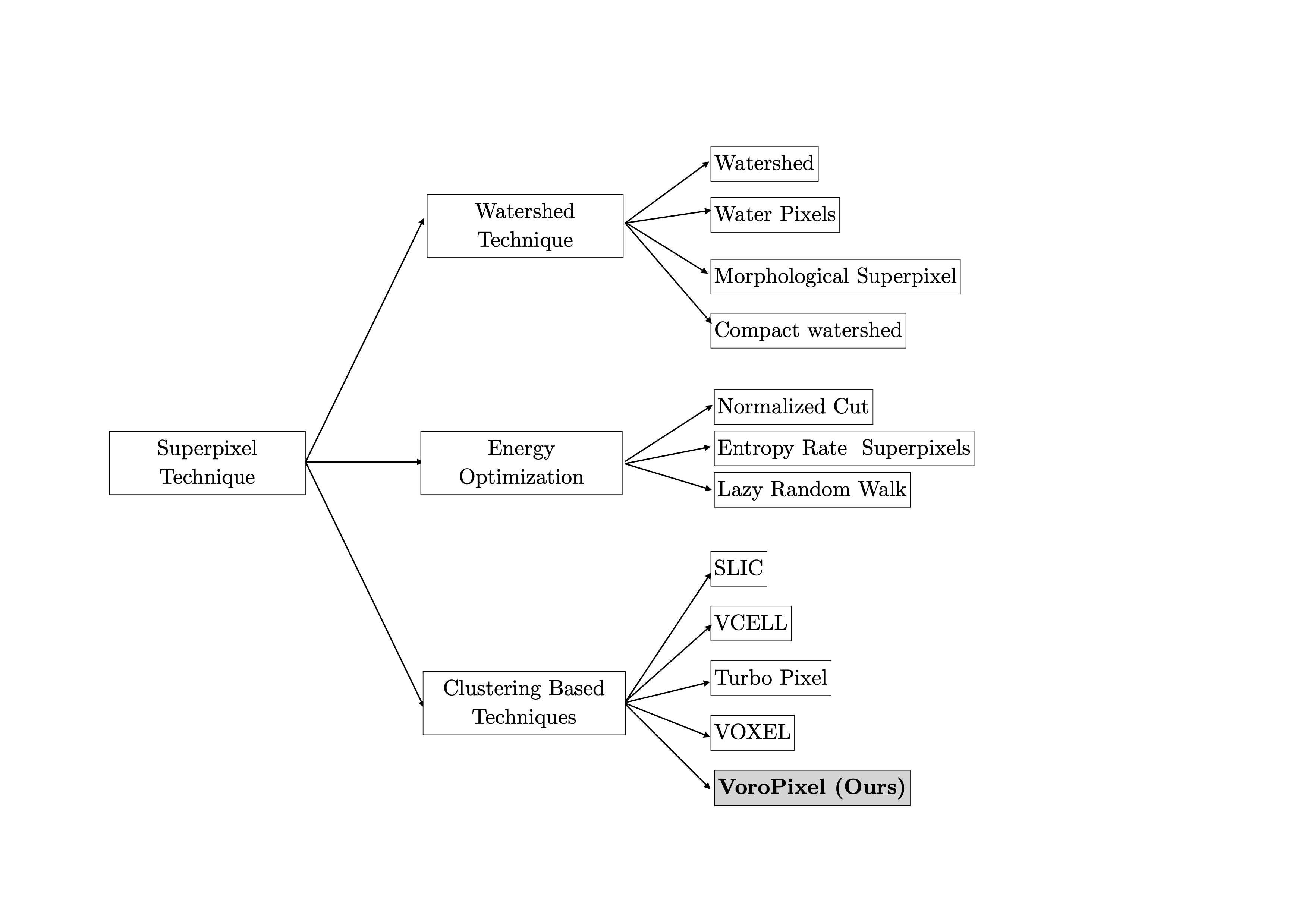}
    \caption{Hierarchical view of different superpixel techniques for images.}
    \label{fig:variation_superpixel}
\end{figure}

An important step in image classification is the integration of GCNs and superpixels. This synergy allows for efficient graph construction, enhanced feature extraction, and improved classification accuracy. Recent studies have demonstrated the effectiveness of this integrated approach in handling high-resolution images and complex visual patterns. 
As many other approaches have been proposed for obtaining superpixels, almost all of them had to preprocess the image for this projection. As such, it yields a significant overhead in the computational complexity of the underlying machine learning task. Various techniques of the superpixels have been studied by \cite{6-kumar2023extensive} in a categorized way. Fig. \ref{fig:variation_superpixel} shows the different ideas in this subject.

To avoid the computational complexity of creating superpixels as an over-head of the preprocessing for the machine learning tasks, a very good and efficient option is to create a Voronoi diagram in just $O(n)$ in which the $n$ is the number of pixels of the image.

In addition, the role of Voronoi diagrams in image classification has taken a significant step in this direction. The Voronoi diagram's capability to create distinct regions based on proximity offers a unique way to segment images, which can complement the graph-based approach of GCNs. Aurenhammer et al. provided a comprehensive review of Voronoi diagrams and their properties \cite{4-aurenhammer1991voronoi}. In the context of image classification, Voronoi diagrams can be used to refine the graph representation of an image further, where each cell in the diagram represents a vertex, enhancing the GCN’s ability to process and classify complex image data.



\section{The Proposed Method: NVGCN}\label{proposed_method} 
In order to describe the main idea of NVGCN, let us first introduce some preliminaries, notation and formalisms.
\subsection{Voronoi Diagrams}
Let 
$\textbf{V}=\{\mathbf{p_1}, \dots,\mathbf{p_m}\subset \mathbb{R}^2 $ where $m\ge 2 $ and $\mathbf{x_i}\neq \mathbf{x_j}$ for $i\neq j, i,j\in I_n=\{1,\dots,n \}$
We call a region given by 
\begin{equation}
    \mathbf{V(p_i)} = \{\mathbf{x} \big| \|\mathbf{x-x_i}\|\leq \|\mathbf{x-x_j}\| for j\neq i, j\in I_n\}    
    \label{eq:planar_voronoi_polygon}
\end{equation}
the planar ordinary Voronoi Polygon associated with $\mathbf{p_i}$ (or the Voronoi polygon of $\mathbf{p_i}$), and the set given by
\begin{equation}
    \mathcal{V}=\{\mathbf{V(p_1)},\dots,\mathbf{V(p_m)}\}
    \label{eq:planar_ordinary_voronoi_diagram}
\end{equation}
the planar ordinary Voronoi diagram generated by $\mathbf{P}$ (or Voronoi diagram of $\mathbf{P}$).
In equation \ref{eq:planar_voronoi_polygon}, $I_n$, $\mathbf{x}$, and $\mathbf{x_i}$ refers to the space, coordinate of a point in the space, and the $generator_i$ respectively.
We call $\mathbf{p_i}$ of $\mathbf{V(p_i)}$ the \textit{generator point}, \textit{generator}, or \textit{centroids} of the $i$th Voronoi polygon, and the set $\mathbf{P}=\{\mathbf{p_1},\dots, \mathbf{p_m}\}$ the \textit{generator set} of the Voronoi diagram $\mathcal{V}$. In some works, the generator points are referred to as a \textit{site}.

\begin{figure}
    \centering
    \includegraphics[width=0.3\textwidth]{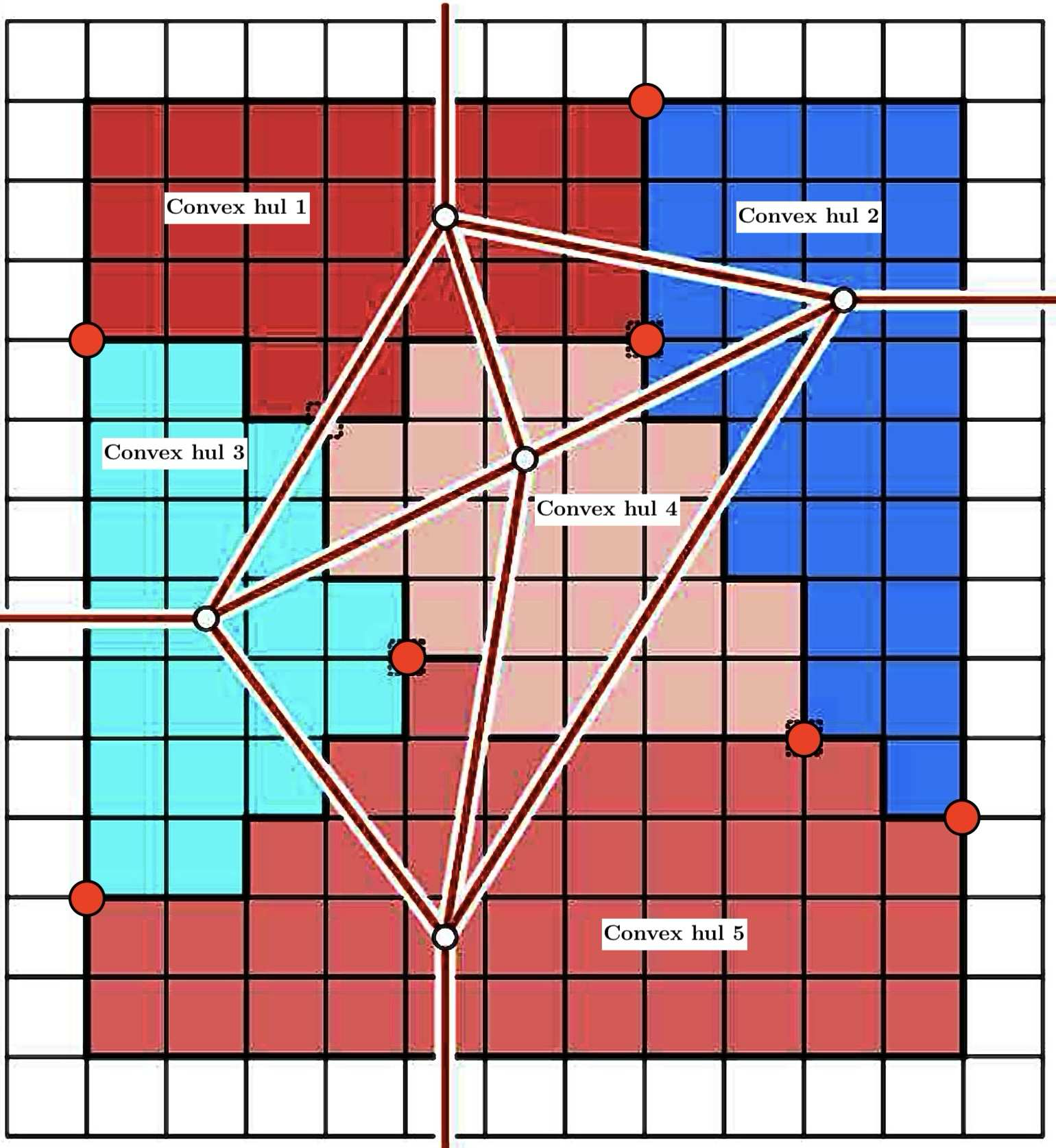}
    \caption{Schematic view of a digital Delaunay triangulation superimposed on the digital Voronoi diagram obtained by flooding.}
    \label{fig:Digitized_vor_tess}
\end{figure}

If $\mathbf{V(p_i)}\cap \mathbf{V(p_j)}\neq \emptyset$, the set $\mathbf{V(p_i)} \cap \mathbf{V(p_j)}$ gives a Voronoi edge. We use $e(\mathbf{p_i}, \mathbf{p_j})$ or $e_{ij}$ instead of $\mathbf{V(p_i)} \cap \mathbf{V(p_j)}$. if $e_{ij}$ is neither empty nor a point, we say that the Voronoi polygons $\mathbf{V(p_i)}$ and $\mathbf{V(p_j)}$ are adjacent.
We can simply change the formula to fit it to the image. The result is a \textit{planar digitized Voronoi diagram} defined as follows:
\begin{equation}
    \textbf{Im}\mathcal{V}=\{\mathbf{\textbf{Im}(V(p_1))},\dots, \mathbf{\textbf{Im}(V(p_m))}\} \, ,
\label{eq:Planar_digitized_voronoi_diagram}
\end{equation}
which \textbf{Im}$\mathbf{V(p_i)}$ is a \textit{digitized Voronoi region}, Im$\delta V(p_i)$ the border of a digitized Voronoi region \textbf{Im}$\mathbf{V(p_i)}$\cite{12-digitized_vor_tess}. It is straightforward to see that in the Digitalized Voronoi diagram $1\le m \le n$. Fig. \ref{fig:Digitized_vor_tess}  shows the digitized Voronoi diagram in the background of the graph.

The proposed method utilizes Voronoi Diagrams and Delaunay Triangulation graphs in the GCN. We name this method Voronoi Graph Convolution Network (VGCN). 


There are two main problems when converting an image into a graph. First, finding out the optimal number of superpixels, which here can be considered as vertices in the graph, is NP-hard. This can be achieved by reducing the problem of finding generator points to the $k$-center problem. We have proved in the appendix of this paper that the optimal number of regions in image segmentation is NP-hard. Also $k$-center problem is investigated more in \cite{kcenterProblem}.  Secondly, even though we may know about the optimal number of superpixels for a particular image, the best centroid's location of these superpixels, the node's features,  is also related to the $k$-center problem by virtue of the fact that to produce the optimal superpixels, the generator points have to be found first. Several techniques have been proposed to solve the $k$-center problem via approximation algorithms \cite{kcenterProblem}. Also, some other techniques resort to clustering approaches via the well-known $k$-means \cite{13-kmeans-lloyd1982least}, \cite{14-kmeans-macqueen1967some}, its variants, as well as other types of clustering algorithms such as hierarchical clustering or fuzzy approaches \cite{HierarchicalClustering_Mittal2022}, \cite{FuzzyClustering_YANG19931}.

For more information regarding the optimal number of regions in an image segmentation, the readers are referred to the appendix.

\subsection{Graph Neural Network Architectures}



Graph Convolutional Neural Networks (GCN) proposed in \cite{GNNModelScarselli2009TheGN} are a class of deep learning models designed to perform inference on data described by graphs. GCNs can capture the dependencies of graphs through messages passing between the vertices of graphs. The fundamental idea behind GCNs is to learn a representation (embedding) for each vertex that captures its own attributes and the collective influence of its neighbors. The representation of a vertex \(v\) is updated iteratively using the following equation:

\begin{equation}
\begin{aligned}
    h_v^{(k)} &= \text{UPDATE}^{(k)} \left( h_v^{(k-1)}, \right. \\
    &\quad \left. \text{AGGREGATE}^{(k)} \left( \{ h_u^{(k-1)} : u \in \mathcal{N}(v) \} \right) \right)
\end{aligned}
\label{eq:GCNFormula}
\end{equation}
\noindent where \(h_v^{(k)}\) is the feature vector of vertex \(v\) at the \(k\)-th iteration, \(\mathcal{N}(v)\) denotes the set of neighbors of \(v\), and UPDATE and AGGREGATE are differentiable functions that update and aggregate features, respectively.

In this context, Message Passing is a mechanism in GCNs that allows vertices to exchange information with their neighbors \cite{18-MessagePassing-Gilmer2017NeuralMP}, facilitating the learning of vertex representations that reflect both their features and the structure of the graph. At each iteration of message passing, a vertex sends a message to its neighbors, which is a function of its current state. The neighbors then aggregate these messages to update their state. The general message-passing rule can be formalized as follows:

\begin{equation}
b_{v}^{(k)} = \sum_{u \in \mathcal{N}(v)} B^{(k)} \left( h_v^{(k-1)}, h_u^{(k-1)}, e_{uv} \right)
\label{eq:MessagePassingFormula}
\end{equation}

\begin{equation}
h_v^{(k)} = U^{(k)} \left( h_v^{(k-1)}, m_{v}^{(k)} \right)
\end{equation}

\noindent where \(b_{v}^{(k)}\) is the aggregated message received by vertex \(v\) in the \(k\)-th iteration, \(B^{(k)}\) and \(U^{(k)}\) are the message and update functions, respectively, and \(e_{uv}\) represents the edge attributes between vertices \(u\) and \(v\).

As a newer generation, Graph Attention Networks (GAT) introduce an attention mechanism into the aggregation function of GCNs \cite{19-GAT-Velickovic2018GraphAN}, allowing vertices to weigh the importance of their neighbors' messages dynamically. This mechanism enables selective focus on information from more relevant neighbors. In GATs, the attention coefficient between two vertices \(u\) and \(v\) is computed as follows:

\begin{equation}
\alpha_{uv} = \frac{\exp\left( \text{LeakyReLU} \left( \mathbf{a}^T [W h_u || W h_v] \right) \right)}{\sum_{k \in \mathcal{N}(v)} \exp\left( \text{LeakyReLU} \left( \mathbf{a}^T [W h_u || W h_k] \right) \right)} \, ,
\label{eq:GAT}
\end{equation}

\begin{figure}
    \centering
    \includegraphics[width=0.5\textwidth]{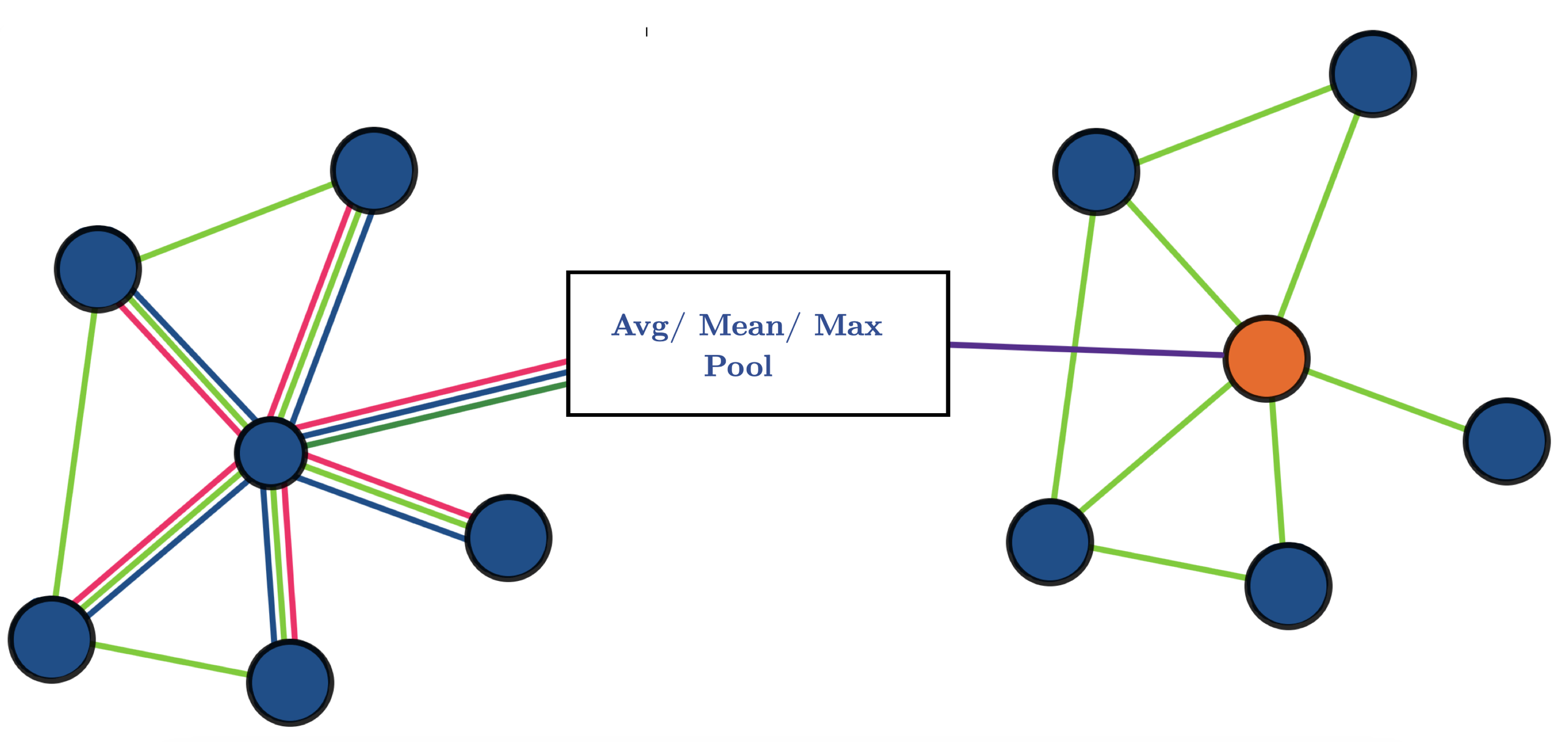}
    \caption{Graph Attention Mechanism. The figure shows how a vertex will be outweighed by another vertex in the attention mechanism. Also, The figure shows the different types of aggregations. Different colors mean different attention in the model. }
    \label{fig:GAT}
\end{figure}

The vertex representation is then updated by weighting the neighbor's features by the attention coefficients:

\begin{equation}
h_v^{\prime} = \sigma \left( \sum_{u \in \mathcal{N}(v)} \alpha_{uv} W h_u \right) \, ,
\end{equation}

\noindent where \(\mathbf{a}\) is a learnable parameter vector, \(W\) is a weight matrix applied to every vertex, \(||\) denotes concatenation, and \(\sigma\) is a non-linear activation function. This attention mechanism enables GATs to adaptively learn the relative weights of neighbors' contributions to a vertex’s new feature representation.
Fig. \ref{fig:GAT} depicts the attention and the aggregation mechanism with regard to the $\alpha$ parameter.

\subsection{The VGCN Architecture}
\begin{figure}[htbp]
\centering
\begin{tikzpicture}[node distance=1.5cm]
    \node [block] (init) {Image};
    \node [block, below of=init] (split) {Split the image into $k$ regular and non-overlapping areas};
    \node [block, below of=split] (snic) {Utilize SNIC to refine the boundaries and best centroids};
    \node [block, below of=snic] (connect) {Connect generator to form a Delaunay triangulation};
    \node [block, below of=connect] (output) {Obtained graph is the Delaunay triangulation};
    
    \path [line] (init) -- (split);
    \path [line] (split) -- (snic);
    \path [line] (snic) -- (connect);
    \path [line] (connect) -- (output);
\end{tikzpicture}
\caption{Process of converting an image into the Delaunay triangulation graph.}
\label{blockdiagram1}
\end{figure}
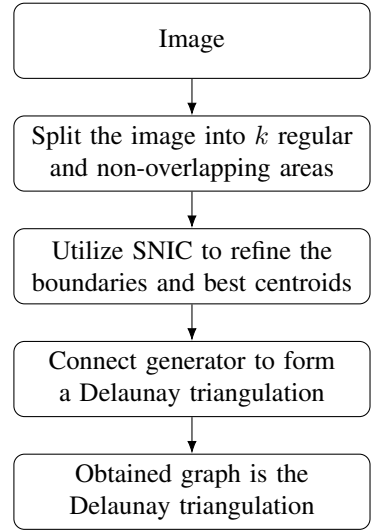

The first step in the pipeline is to convert the input image into the corresponding Delaunay triangulation graph. The block diagram of Fig. \ref{blockdiagram1} shows the main steps for this conversion. Algorithm {\em Delaunay Triangulation} depicts the pseudocode for this step in more detail.

\begin{algorithm}
\caption{Delaunay Triangulation (Algorithm III-C)} 
  \textbf{Input:} image, $\mathbf{Img}$, number of expected Delaunay triangulation graph's vertices, $m$\\
  \textbf{Output:} Delaunay triangulation graph, $G_{Tess}$\\
    \begin{algorithmic}[1]
    \State ${L, S, P are label map, segments and the centroids respectively}$    
    \State{$\mathbf{L},\mathbf{S} , \mathbf{P} \gets \text{SNIC($\mathbf{Img}$, $m$)}$}
    \State{$\textbf{X}, \mathbf{E} \gets 
    \text{
    Douglas\_Peucker($\mathbf{S}$, $e$)}
    $}
    \State{$\mathbf{E_{Tess}}\gets \{\emptyset\}$}
    \For{$x_1, x_2 \in E$}
        \State{$r_1, r_2\gets  \mathbf{L}[\mathcal{N}_{x_1}] \cap \mathbf{L}[\mathcal{N}_{x_2}]$} 
        \State{\#Finding the generator points of the shared regions}
        \State{$p_1, p_2 \gets P[r_1], P[r_2]$}
        \If{$(p_1, p_2) \perp (x_1, x_2)$}
            \State{$\mathbf{E_{Tess}} \gets \mathbf{E_{Tess}} \cup (p_1, p_2)$}
        \EndIf
    \EndFor
    \State{$\mathbf{G_{Tess}} \gets (\mathbf{P, E_{Tess}})$}
    \State{\textbf{Return} $G_{Tess}$}
  \end{algorithmic}
  \label{alg:algorithmDelaunay}
\end{algorithm}

In the algorithm \ref{alg:algorithmDelaunay}, $\mathbf{L, S, P}$ are the label or region map, SNIC boundaries and the generator points sets, respectively. $(\mathbf{X, E})$ can be interpreted as the Voronoi diagram graph with regards to the regions $\mathbf{V}(p_i)| \forall 1\leq i \leq v$ \textit{as Voronoi vertices set}, while $\mathbf{E}$ is the set of Voronoi edges. Also,  $r_1, r_2$ refers to the label map of the two shared regions in a particular Voronoi edge $e$. Finally, $p_{r_1}, p_{r_2}$ represents the generator points of the regions $r_1$ and $r_2$, respectively. In the remaining of this paper, we call these regions Voronopixels. The for loop implements the creation of Tessellation edges as follows:
\begin{equation}\label{eq:tessEdgeBuilder}
    \begin{split}
        E_{Tess} &\gets\\
        \{(p_i,p_j)\} &\iff \exists (p_i, p_j)  \in \mathbf{P}, \\
        &(p_i, p_j) \perp (x_l, x_k)\in \mathbf{X} ,\\
        &\text{ region\_id}_{neighbors_{x_l}= \text{region\_id}}\{p_i, p_j\} 
    \end{split}   
\end{equation} \,

In this equation, $(p_i, p_j)$ is an edge in the Delaunay triangulation, $p_i$ is the generator point of the region defined by SNIC, $(x_l,x_k)$ is an edge in the Voronoi Diagram generated using the Douglas Peucker algorithm, $x_l$ and $x_k$ are the Voronoi vertices which is the boundaries intersection with at least shared three regions. Not to mention that $|P|=m$ and $|V|=v$. $P$ and $V$ are the Voronoi and the Delaunay triangulation vertices, respectively.\\

The main process is depicted in the block diagram of Fig. \ref{blockdiagram1}, and the algorithm is depicted in Algorithm \ref{alg:algorithmDelaunay}. We used SNIC to obtain the superpixels \cite{15-SNIC-achanta2017superpixels}. It is a very fast convergence and memory-efficient method for coarsening an image into some regions of superpixels compared to SLIC and LSC \cite{11-LinearSpectralClustering7814265}. 

From a computational complexity standpoint, SNIC is not an iterative algorithm. The time complexity of this algorithm is $O(n)$, where $n$ is the number of pixels. It does not use $k$-means clustering to create superpixels. Instead, by using the Douglas-Peuker algorithm \cite{16-DouglasPeuker}, we refine the boundaries of the superpixels from non-straight line edges to straight line edges. The algorithm uses $\alpha$ as a hyperparameter to adjust the edges. Since the number of vertices in the Voronoi diagram and the corresponding Delaunay triangulation graph is at most $n$, the running time of this algorithm depends on the number of vertices and their degrees in the graph, namely, $O(|X|.|E|)$, where $|X|.|E| \ll n$. The for loop and its nested statements also run in $O(|E|)$. As a result, the entire algorithm runs in $O(n)$.

\begin{figure}
    \centering
    \includegraphics[width=0.3\textwidth]{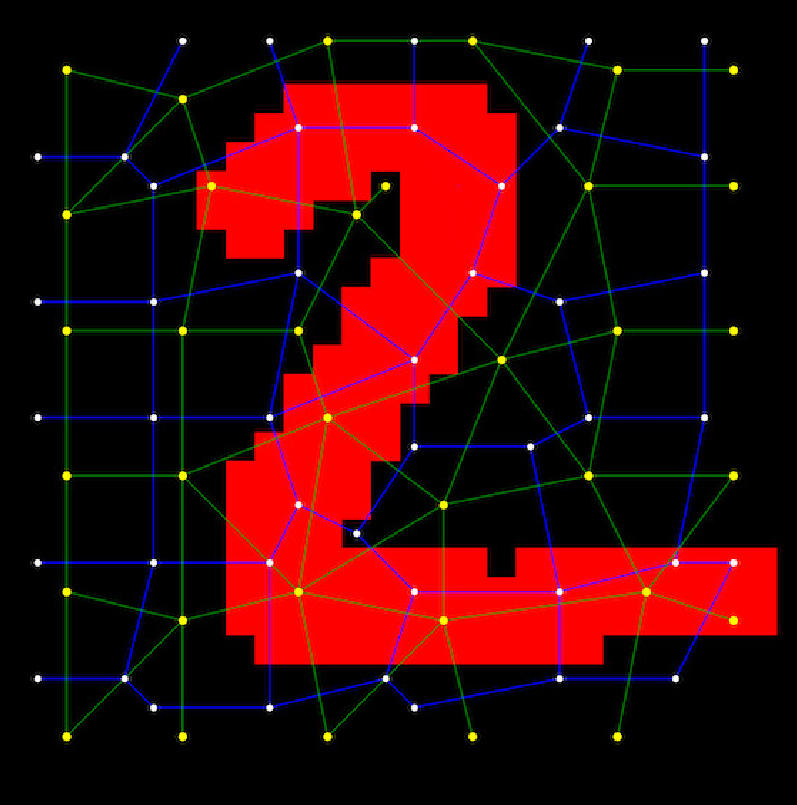}
    \caption{Voronoi diagrams (the blue graph) and the related Delaunay triangulation (the green graph) on an image of MNIST dataset created by the suggested method.}
    \label{fig:Delaunay_triangulation}
\end{figure}

Thus, by running Algorithm \ref{alg:algorithmDelaunay}, we obtain both the Voronoi diagram and the corresponding Delaunay triangulation graph in $O(n)$. Just as a recap, a Voronoi Diagram and the Delaunay triangulation follow the duality relationship.
Fig. \ref{fig:Delaunay_triangulation} shows this duality relationship. In the figure, the Voronoi diagram and its triangulation are shown as blue and green graphs, respectively.

\begin{figure}
    \centering
    \includegraphics[width=0.24\textwidth]{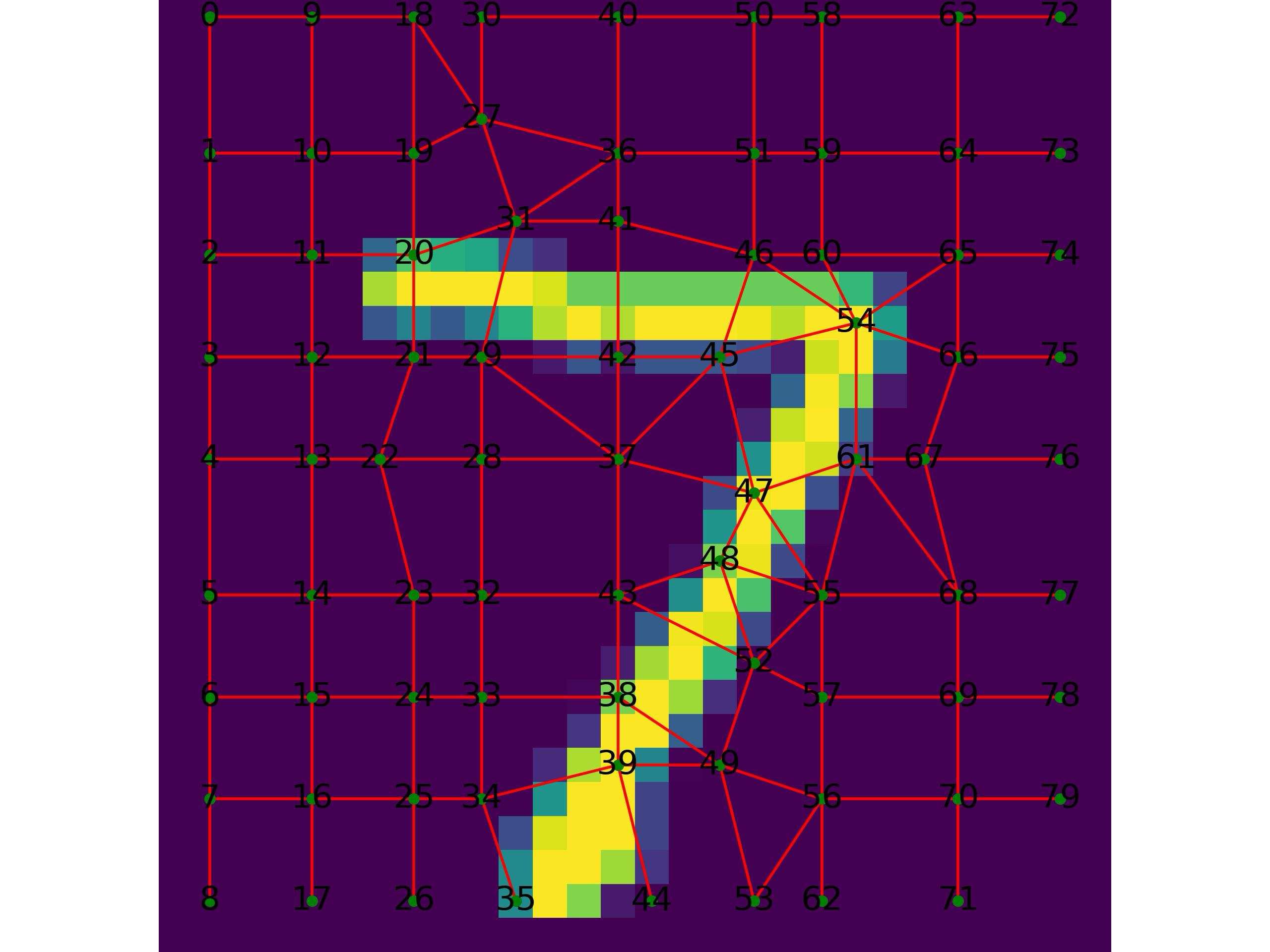}\hfill
    \includegraphics[width=0.24\textwidth]{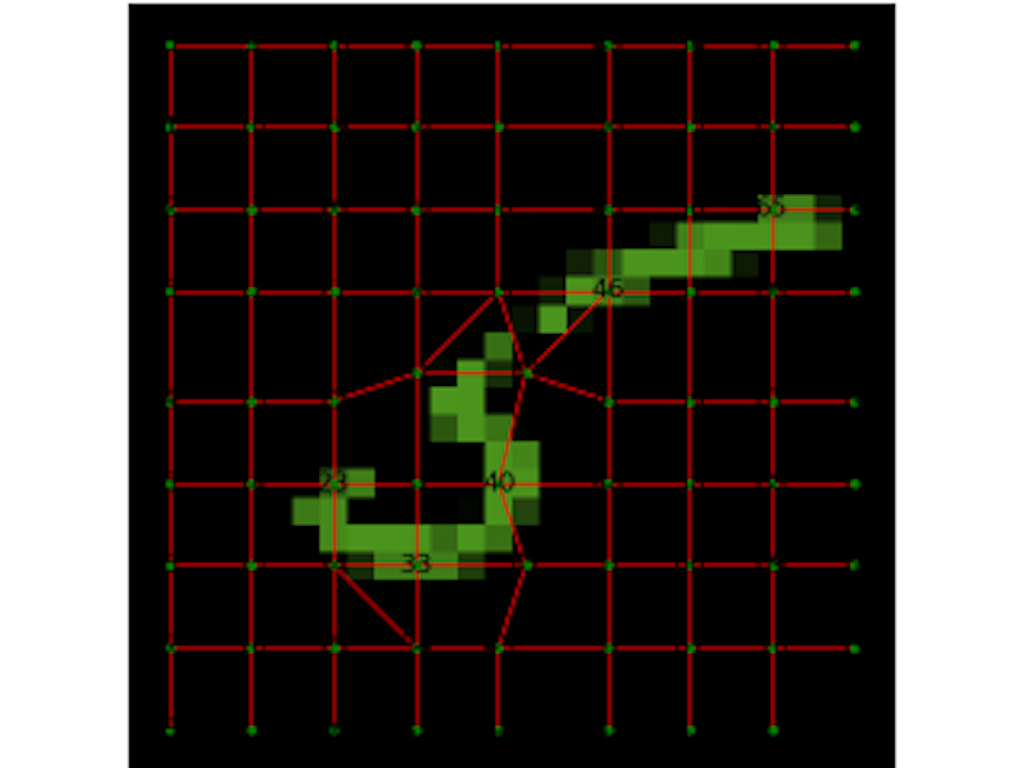}\hfill
    \\[\smallskipamount]
    \includegraphics[width=0.24\textwidth]{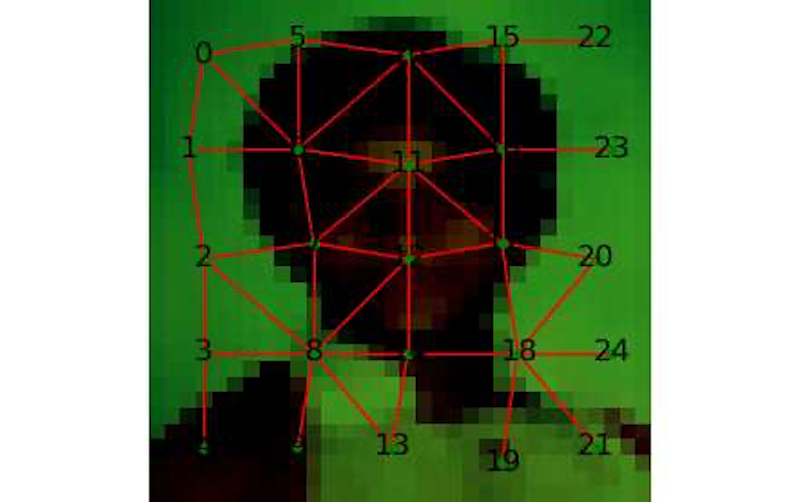}\hfill
    \includegraphics[width=0.24\textwidth]{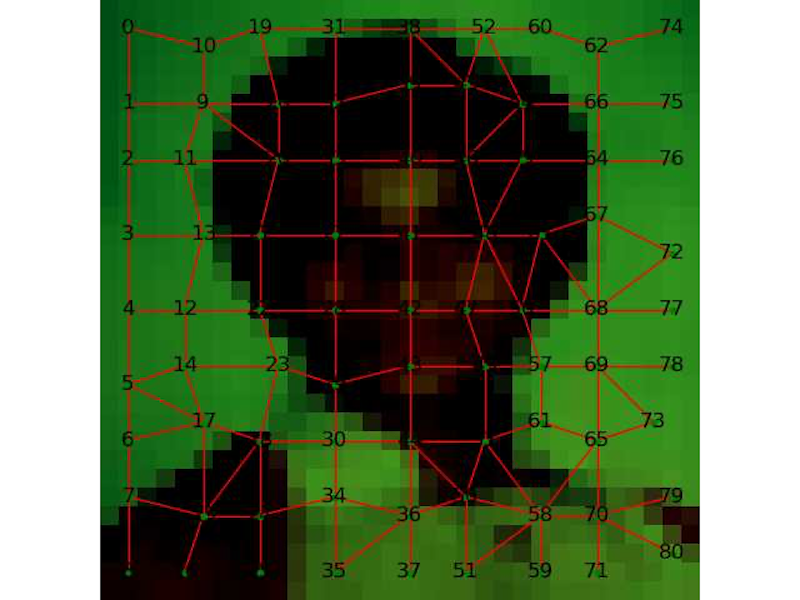}\hfill
    \caption{Delaunay triangulation obtained with approximately 81 generator points for the MNIST and CIFAR100 datasets. As shown in the image, we observe that the number of edges around the edge of the digits is deformed.}
    \label{fig:Tessellation_graph_128}
\end{figure}

Also, in Fig. \ref{fig:Tessellation_graph_128}, the output of the proposed algorithm is illustrated. We observe that the resulting graph is more sensitive around the edges. The number of edges for each vertex around the edges is more likely to be higher, and the topology of the connected vertices around the edges is more deformed than other vertices away from the edges. From an experimental point of view, the vertices spread out in a wide range of backgrounds without major changes and are more likely to be a kind of regular grid than other vertices with different regions' color intensity.

\subsection{Normalized Voronoi Graph Convolution Network (NVGCN)}

Graph Convolutional Networks (GCNs) primarily leverage the degree matrix for normalization. In graph data, nodes can have widely varying degrees, meaning some nodes are connected to many neighbors, while others are connected to very few. Without normalization, nodes with many connections (high-degree nodes) can disproportionately influence feature propagation, while low-degree nodes may be underrepresented. To address this, GCNs incorporate the degree matrix to normalize the adjacency matrix. The degree matrix \(D\), a diagonal matrix where each entry corresponds to the degree of a node, is used to normalize the adjacency matrix \(A\) in a symmetric way as \( \tilde{A} = D^{-\frac{1}{2}} A D^{-\frac{1}{2}} \). This ensures that feature aggregation from neighbors is balanced, giving fair representation to both high-degree and low-degree nodes.

This approach was introduced by Kipf and Welling (2017) in their seminal work on GCNs \cite{kipf2017semi}. By using this normalization technique, GCNs avoid the issue of information from high-degree nodes dominating the learning process, which could lead to unstable updates. Normalization by using the degree matrix allows for more stable and effective learning, ensuring that each node contributes appropriately to its neighbors' feature updates, regardless of its degree. The method has become a standard in the field of graph neural networks for handling irregular data structures in graph-structured data.

This means that if the graph used for the GNN is already normalized,  the degree matrix $D$ in the main formula can be removed.  The Delaunay triangulation graph (D.T graph) is the one we use for this purpose because the degree $k$ in the graph will not exceed the value of 8 in total.  The degree in a D.T graph is 6 by approximation \cite{book_deBerg2008}.  By knowing this fact, not only we can remove the degree matrix in the base formula of the GCN,  but the time complexity of computation, the number of multiplication operation, and the aggregation information time from the local nodes (neighbors) will decrease drastically.

\begin{figure}[ht!]
\centering
\includegraphics[width=0.8\linewidth]{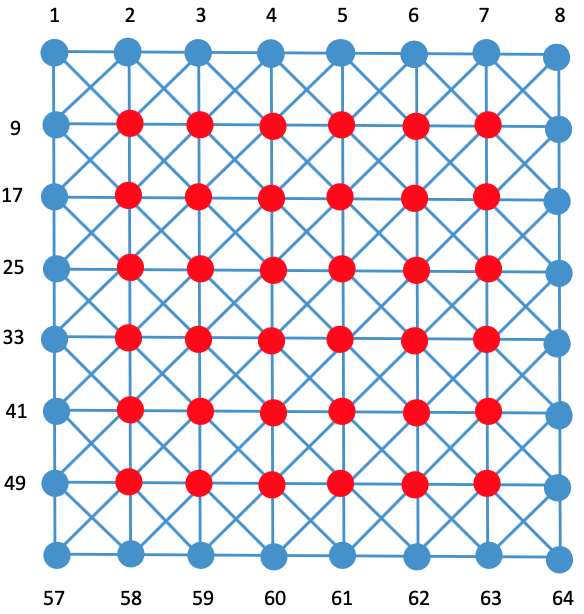}
\caption{The densest Delaunay triangulation in a graph. The densest graph is that each pixel acts as a node. The blue nodes are the edges of the image compared to the red nodes with the highest degree in the D.T graph.}
\label{FigdensestDelaunay}
\end{figure}

In particular, the densest D.T graph created from an image is shown in Fig. \ref{FigdensestDelaunay}. The densest D.T graph in an image occurs when each pixel acts as a node in the image.  In this example, the degree of nodes 1, 8, 57,  and 64 is three.  All other nodes' degrees on the edge of the image are equal to five.  Also, the highest degree in this graph is related to the nodes within the graph, which is eight and is shown in red. 

\begin{figure}
    \centering
    \includegraphics[width=\linewidth]{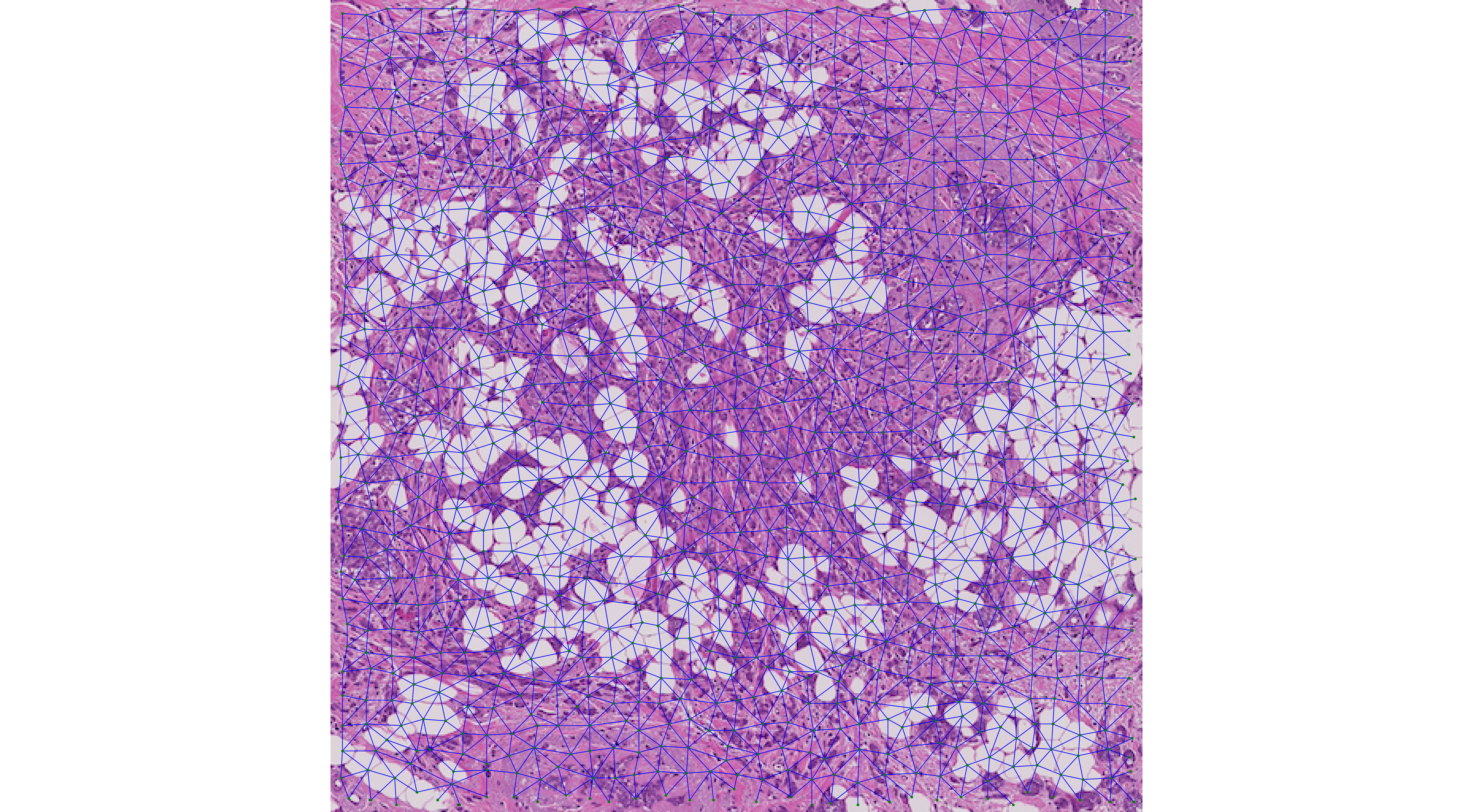}
    \caption{A Bio-medical high-resolution image (7000 X 7000 pixels) which using our technique a graph created on top of the image.}
    \label{fig:regionized_TMA_img}
\end{figure}

\begin{figure}
    \centering
    \includegraphics[width=\linewidth]{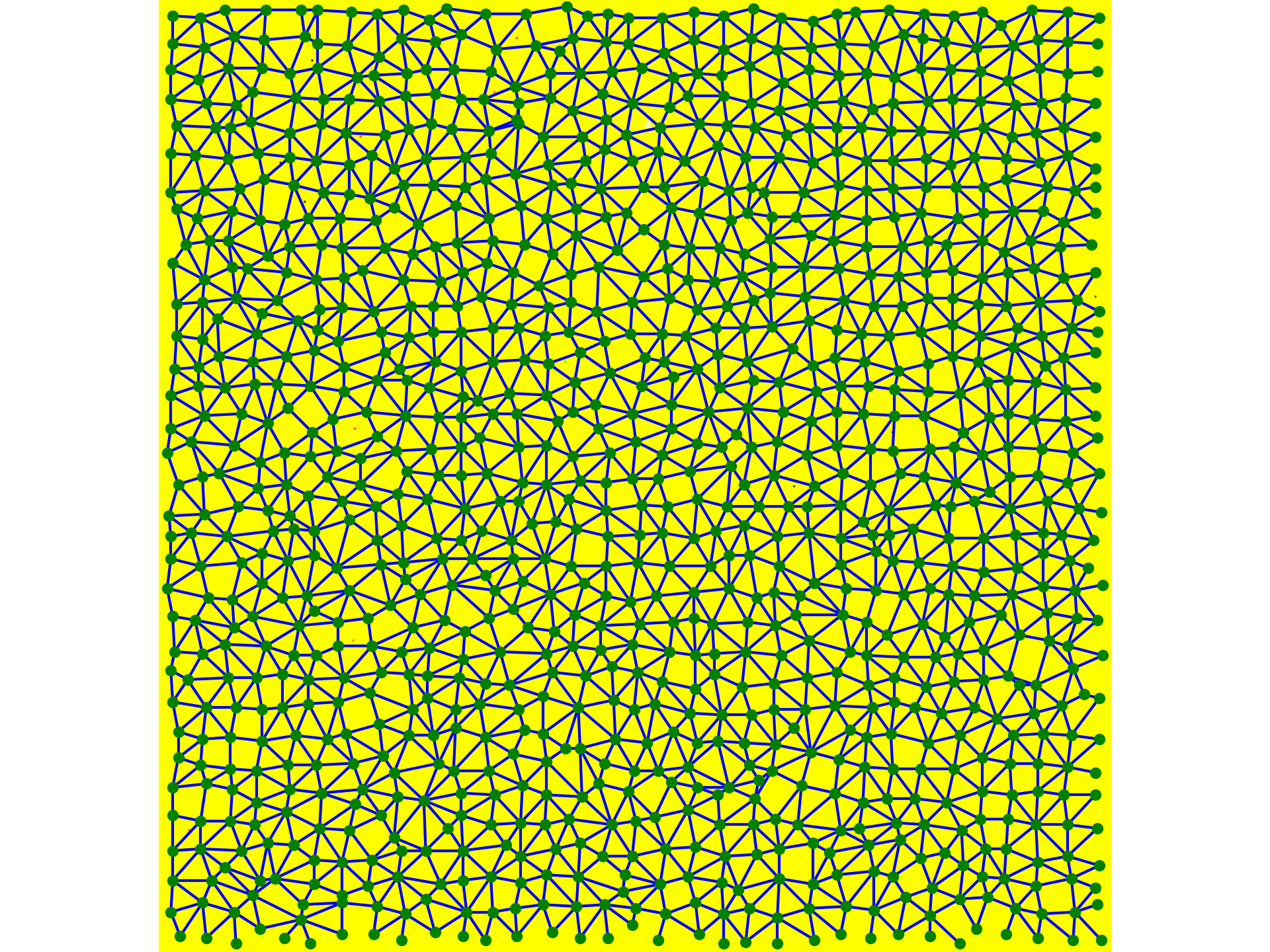}
    \caption{The pure Delaunay graph of a Tissu microarray image.}
    \label{fig:voronoi_TMA_img}
\end{figure}

\begin{figure}
    \centering
    \includegraphics[width=0.8\linewidth]{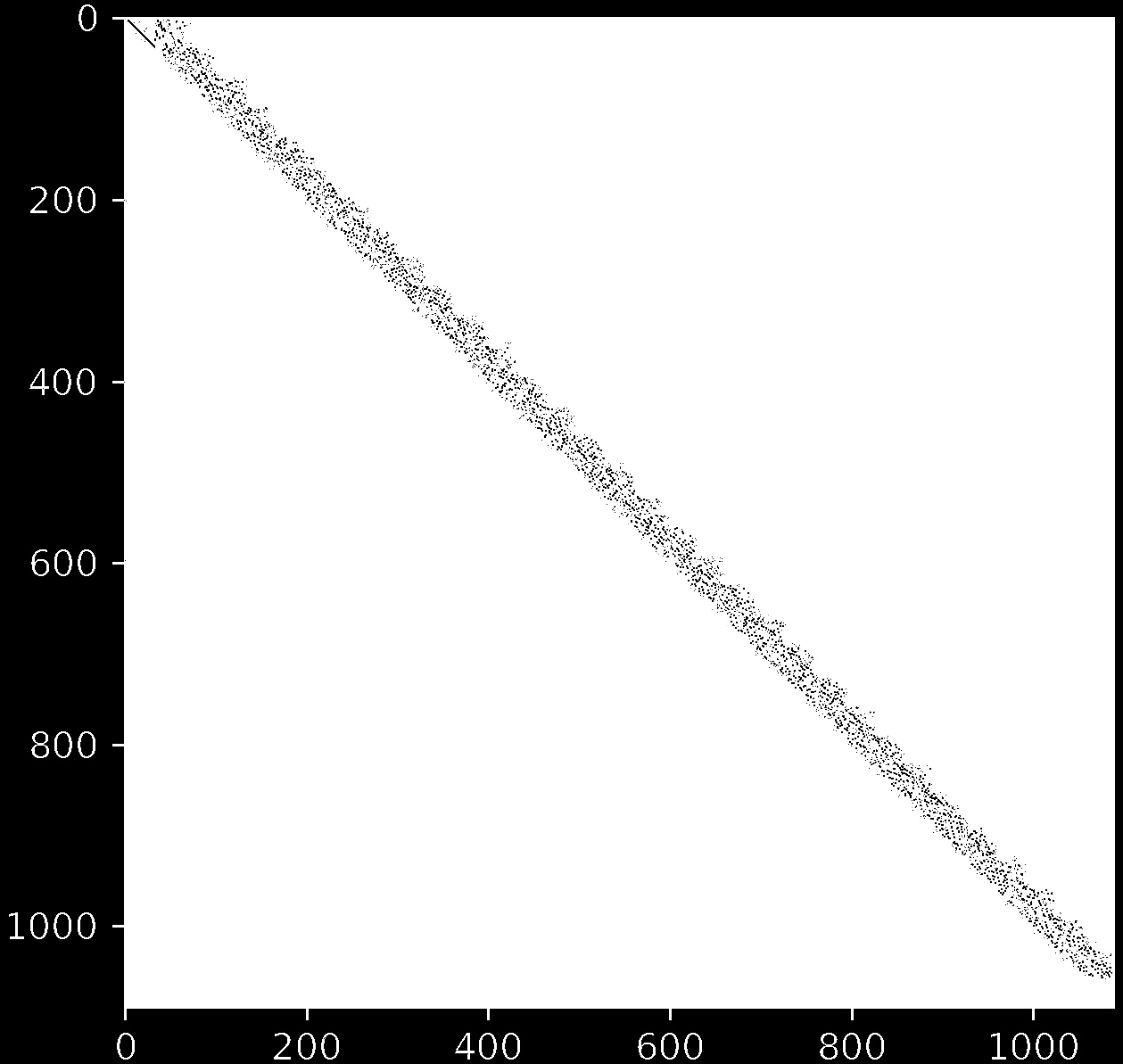}
    \caption{The adjacent matrix shows the sparsity and the proximity of the nodes connected to each other for the TMA's Delaunay graph.}
    \label{fig:adj_TMA_img}
\end{figure}

Fig. \ref{fig:regionized_TMA_img} illustrates a high-resolution Tissue Microarray(TMA) image with [7000 x 7000] pixels which the Delaunay Triangulation graph is plotted on the top of it. The number of nodes in this example is about 1024 nodes. Fig. \ref{fig:voronoi_TMA_img} is the solely Delaunay triangulation graph plotted without the image for clarity. Also, Fig. \ref{fig:adj_TMA_img} shows the adjacency matrix for the related graph from this high-resolution TMA image. As the adjacency matrix plotted in this figure, almost all the nodes with proximity of no more than 10 are connected to each other. Furthermore, the average degree of each node in this particular graph is 6.25, which suggests that using a normalized graph is the best choice.

Thus, instead of applying the degree matrix on the based GCN formula, we can modify the adjacency matrix in such a way that, instead of adding just self-loops, identity matrix, in the diagonal adjacency matrix, we can fill it with the number 6. Thus, the base formula of the regular GCN can be changed as follows:

\begin{align*}
\text{Base Formula:} \quad H^{(l+1)} &= \sigma \left(D^{-\frac{1}{2}} \tilde{A} D^{-\frac{1}{2}} H^{(l)} W^{(l)} \right) \\
\text{Updated Formula:} \quad H_{\text{D.T}}^{(l+1)} &= \sigma \left( \hat{A}_{\text{D.T}} H_{\text{D.T}}^{(l)} W \right)
\end{align*}

\noindent where $H_{\text{D.T}}^{(l+1)}$, $H_{\text{D.T}}^{(l)}$, $H^{(l+1)}$, and $H^{(l)}$ are the feature maps on layer $(l+1)$ and layer $(l)$ in the updated and the base formula, respectively.

\[
\tilde{A} = A + \text{I}
\]
\noindent where $I$ is the identity matrix for adding self-loops.

\[
\hat{A}_{\text{D.T}} = A + (6\text{I})
\]

$\sigma$ is the non-linear activation function, and $W$ is the linear weight matrix, which is trainable, and $6$ in this formula is constant and can be removed as well.

We have provided extensive research on the calculation of the computational complexity and the number of reductions in the multiplication operation in NVGCN layers is in the appendix. The readers can find in the appendix that the average deduction on the number of multiplication is about 33\% to 66\% in the updated formula (NVGCN).

\subsection{Datasets}\label{datasets}
For the current study, we have used CIFAR10 and MNIST datasets, which are the most widely-used benchmarks for image classification. MNIST has 60,000 images with 28 X 28 pixels and 10 different labels. CIFAR-10 has 60,000 different images with 10 different labels. 
We split each dataset into 80\% and 20\% as the training and evaluation dataset samples, respectively. A summary of these three datasets is shown in the table \ref{table:dataset_comparison}.

\begin{table}[h]
    \centering
    \caption{Comparison of CIFAR-10, Fashion-MNIST, and MNIST datasets.}
    \label{table:dataset_comparison}
    \begin{tabular}{|l|c|c|c|c|}
        \hline
        \textbf{Dataset} & \textbf{\# Images} & \textbf{\# Labels} & \textbf{Image size} & \textbf{Type} \\ \hline
        CIFAR-10 & 60,000 & 10 & 32x32 & RGB \\ \hline
        FashMNIST & 60,000 & 10 & 28x28 & Gray \\ \hline
        MNIST & 60,000 & 10 & 28x28 & Gray \\ \hline
    \end{tabular}%
\end{table}

\section{Experiments}

First of all, in terms of converting an input image into the Delaunay triangulation graph, we set the number of $k=64$ as the regular number of generator points for the $28\times 28$ and $32 \times 32$ images. For the consistency of the created graph and to capture the geolocation of the generator points, we set $S$ in Equation (\ref{eq:SLIC_formula}) to 50. In this way, we ensure that the location of the generator points will be maintained. The number of epochs for all experiments was set to 300 with an early stopping of 10 epochs. Early stopping has been used to ensure that the over-smoothing and overfitting are minimized. The learning rate in our experiments was set to 0.001, and remains fixed across all experiments. For all batches, the batch size is set to 128 graphs. The number of inputs is set to 3, which is the location of the generator points and the means of the normalized colour intensity for each convex hull in the Delaunay triangulation. The training and test size ratio is 0.2, which means 20\% of the dataset is used as a test set.

We have used three layers of the GCN for the MNIST and FashionMNIST datasets as a general framework utilizing multi-head Graph Attention Networks with three heads. The input of the model is the set of Delaunay triangulation vertices and the set of edges, $G_{Tess} = (\mathbf{P}, E_{Tess})$.
After applying three layers of multi-head attention layers, we flatten it with the global mean pooling and utilize two more MLP layers before the $y_{hat}$ layer for the classification.

The average number of Voronoi vertices and edges for this dataset is approximately 72 and 165 for each image, respectively. We have used Pytorch-Geometric to handle the graph and for utilizing the GAT architecture \cite{20-pygeometric-Fey2019PyTorch}. A summary of the model is shown in Table \ref{tab:GATMultiHead}.

\begin{table}[h!]
\centering
\caption{GATMultiHead Model Structure}
\begin{tabularx}{\linewidth}{|>{\centering\arraybackslash}X|>{\centering\arraybackslash}X|>{\centering\arraybackslash}X|>{\centering\arraybackslash}X|}
\hline
\textbf{Layer} & \textbf{Input Features} & \textbf{Output Features} & \textbf{Details} \\ \hline
GATConv & 3 & 32 & head=2 \\ \hline
BatchNorm & N/A & 64 & N/A \\ \hline
GATConv & 64 & 64 & head=2 \\ \hline
BatchNorm & N/A & 128 & N/A \\ \hline
GATConv & 128 & 64 & head=2 \\ \hline
BatchNorm & N/A & 128 & N/A \\ \hline
Linear & 128 & 32 & bias=True \\ \hline
BatchNorm & N/A & 32 & N/A \\ \hline
Linear & 32 & 32 & bias=True \\ \hline
Linear & 32 & 10 & bias=True \\ \hline
\end{tabularx}
\label{tab:GATMultiHead}
\end{table}

The input features for the vertices are 5, which is the color intensity with the vertex's coordinate.

\noindent
\begin{equation}
    H(X) = -\sum p(X)\log p(X) \, ,
    \label{entropy}
\end{equation}

\section{Performance Evaluation and Results}

We conducted extensive experiments to evaluate the performance of our proposed method with other approaches. We included the following methods for comparison: Image Classiﬁcation using Graph Neural Network and Multiscale Wavelet Superpixels (Wave Mesh) \cite{Wavelet_vasudevan2023image}, Superpixel Image Classification with Graph Attention Networks (HGNN)\cite{RAG_avelar2020superpixel}, Spatial Graph Convolutional Networks (SGCN), and Topological Graph Neural Networks GNN algorithms (TOGL). Table \ref{tab:combined_gnn_results} combines the best-reported results from our implementations for the above methods, compared to those published in the original papers, organized by datasets. SP is used as a shorthand for superpixel to make the table more visually appealing. The best result achieved by our method for each dataset is highlighted in bold, along with the best result reported by each author's implementation for each dataset. We reported our method (VGCN) results in this table as well. Grid is listed next to certain method entries for additional context to their graph representation, and "PC" or "Mac" is used in each case for context as to why there is a duplicate entry.

\begin{table}[!htbp]
\centering
\caption{\ centring Comparison of our implementation of three GNN model performances on each dataset to the author's reported results.}
\footnotesize
\begin{tabular}{>{\raggedright\arraybackslash}p{1.5cm} >{\raggedright\arraybackslash}p{2cm} >{\centering\arraybackslash}p{2cm} >{\centering\arraybackslash}p{2cm}}
\toprule
\textbf{Dataset} & \textbf{Method} & \textbf{Authors' Results (\%)} & \textbf{Our Results (\%)} \\ 
\midrule
\multirow{8}{*}{MNIST} 
 & HGNN-2Head & 96.19 & \textbf{96.22} \\
 & SGCN (Grid) & \textbf{99.61} & -- \\
 & SGCN (SP) & 95.95 & 93.07 (PC) \\
 & SGCN (SP) & 95.95 & 89.76 (Mac) \\
 & GCN-4 & 90.0$\pm$0.3 & 77.25 \\
 & GCN3-TOGL-1 & 95.5$\pm$0.2 & 90.60 \\
 & \textbf{VGCN (ours)} & -- & \underline{\textbf{95.5}}\\
\cmidrule{1-4}
\multirow{7}{*}{FashionMNIST} 
 & HGNN-1Head & \textbf{83.07} & \textbf{82.95} \\
 & HGNN-4Head & -- & 74.86 \\
 & HGNN (Grid) & -- & 78.89 \\
 & GCN-4 & -- & 64.31 \\
 & GCN3-TOGL-1 & -- & 76.95 \\
 & \textbf{VGCN (ours)} & -- & \underline{\textbf{81.4}}\\
\cmidrule{1-4}
\multirow{6}{*}{CIFAR-10} 
 & HGNN-2Head & 45.93 & \textbf{66.08} \\
 & SGCN (Grid) & -- & 49.27 \\
 & GCN-4 & 54.2$\pm$1.5 & 47.38 \\
 & GCN3-TOGL-1 & \textbf{61.7$\pm$1.0} & 37.41 \\
& \textbf{VGCN (ours)} & -- & \underline{\textbf{45}}\\
\bottomrule
\end{tabular}
\label{tab:combined_gnn_results}
\end{table}


To achieve a fair comparison, all accuracies are reported based on the number of vertices. We converted each image into the corresponding Delaunay triangulation with 64-85 vertices for all three datasets. Also, we did not fine-tune the model's hyper-parameters to achieve the best accuracy level.


For the MNIST dataset, our VGCN model achieved an accuracy of 95.47\%. This result is particularly notable given the simplicity of the dataset and the baseline accuracy of other GCNs on this dataset, which typically ranges between 85-97\%. The slight drop in accuracy compared to GCNs is offset by our method's increased efficiency and speed, making it a viable option for scenarios where computational resources are limited.


The Fashion MNIST dataset, which consists of 60,000 training images and 10,000 testing images of clothing items, presents a more complex classification task than the MNIST dataset due to higher variability in the visual features of the items. Our VGCN model achieved an accuracy of 81.4\%. The ability of VGCN to effectively capture and process relational and contextual information in the images contributed significantly to its performance on this dataset.


The CIFAR-10 dataset is considerably more challenging due to the images' higher resolution and greater diversity. Our VGCN model achieved an accuracy of 45\%. The relatively lower performance suggests that further optimization and tuning of the VGCN model are required to leverage its capabilities on such datasets fully.

\subsection{Computational Efficiency}

One of the key advantages of our VGCN model is its computational efficiency. Using Voronoi diagrams for graph construction significantly reduces the preprocessing time compared to traditional superpixel methods like SLIC. The overall time complexity for converting an image into a graph using our method is $O(n)$, where $n$ is the number of pixels in the image. This efficiency is crucial for applications requiring real-time image processing and classification. Also, in our work, the speed of the GCN is much faster than that of traditional GCNs. This is because of the sparsity of the Adjacency matrix for the Delaunay triangulation graph. The average number of edges for each vertex is approximately $6.25$, indicating that the graph is sparse compared to all other methods.

\textbf{Robustness}

In addition, the robustness of the VGCN model was tested by introducing variations in the image data, such as noise and occlusions. The model maintained a high level of performance under these conditions, demonstrating its ability to generalize well to different types of image distortions. This robustness can be attributed to the graph-based representation of images, which captures both local and global features more effectively than traditional grid-based representations.

In terms of preprocessing, to convert an image to the corresponding graph, we tracked the time for each image in a single and multi-processing manner. Converting an image into a graph just took 0.023 seconds. We used 12 core CPUs to convert the MNIST and FashionMNIST images to graphs, and it took 110 seconds to convert 60,000 images. The baseline model was a bit faster, around 10 seconds. The reason behind that is that converting the image into a Delaunay triangulation needed a few more computations than SNIC. On the other hand, the VGCN algorithm took less time to converge.

Finally, the results of our experiments highlight several key points regarding the performance and applicability of the VGCN model:

- Efficiency in Graph Construction: Using Voronoi diagrams for image partitioning and graph construction efficiently and effectively captures the relational context of image regions. This efficiency makes the VGCN model suitable for real-time applications. The number of regions/superpixels is shown in table \ref{table:vertices_comparison}. As it shown in the table, we found that when an image is more complex, the number of regions, nodes in the graphs, should be more.

- Competitiveness with Traditional Methods: On simpler datasets like MNIST and Fashion MNIST, the VGCN model demonstrated competitive performance with regular GCNs. Although the accuracy on the more complex CIFAR-10 dataset was lower, the model's potential for improvement through further optimization is evident.

- Robustness to Variations: The robustness of the VGCN model to noise and occlusions is a significant advantage, making it a strong candidate for applications in dynamic and unpredictable environments.

After running our method and other state-of-the-art methods, the results are summarized in Table \ref{tab:combined_gnn_results}.

\begin{table}[h]
\centering
\caption{Number of vertices used by different models on various datasets.}
\begin{tabularx}{\linewidth}{|>{\centering\arraybackslash}X|>{\centering\arraybackslash}X|>{\centering\arraybackslash}X|>{\centering\arraybackslash}X|}
    \hline
    \textbf{Model} & \textbf{MNIST vertices} & \textbf{Fashion MNIST vertices} & \textbf{CIFAR-10 vertices} \\ \hline
VGCN & 64 & 64 & 150 \\ \hline
Superpixel GAT & 75 &75 & 150 \\ \hline
\end{tabularx}
\label{table:vertices_comparison}
\end{table}

In addition to these comparisons, we have used three different machine learning architecture, GCN, GAT, and NVGCN for tracking the times spent for each epoch. All of these three architectures use the graphs created based on the Voronoi diagram and the Delaunay Triangulation which is discussed earlier in the paper. We changed the first architecture as VGCN, Voronoi Graph Convolution Network which is used for both GCN and GAT architectures. The third architecture, NVGCN, Normalized Voronoi Graph Convolution Network, is the VGCN with the degree matrix out of the original GCN formula. We used this architecture for analyzing the VGCN  and GAT in terms of speed and accuracy. We used the same setting for the model for NVGCN as the table \ref{tab:GATMultiHead} for fairness. Also, table \ref{tab:time_comp_NVGCN} show the time and the accuracy percentage of the VGCN and NVGCN.

\begin{table}[h]
    \centering
	\begin{tikzpicture}
		\begin{axis}[
			xlabel={Epoch},
			ylabel={Time (minutes)},
			legend style={font=\tiny, at={(1,1)}, anchor=south east},
			grid=both,
			ymin=0, ymax=0.6,
			xtick={1,2,3,4,5,6,7,8,9,10},
			ytick={0,0.1,0.18,0.2,0.3,0.4,0.5,0.6}
			]
			
			\addplot[color=blue,mark=*] coordinates {
				(1, 0.19) (2, 0.19) (3, 0.20) (4, 0.20) (5, 0.21)
				(6, 0.21) (7, 0.21) (8, 0.20) (9, 0.20) (10, 0.20)
			};
			\addlegendentry{Regular GCN}
			
			\addplot[color=red,mark=square*] coordinates {
				(1, 0.16) (2, 0.18) (3, 0.17) (4, 0.18) (5, 0.18)
				(6, 0.18) (7, 0.18) (8, 0.18) (9, 0.18) (10, 0.18)
			};
			\addlegendentry{NVGCN\_pyg}
			
			\addplot[color=green,mark=triangle*] coordinates {
				(1, 0.51) (2, 0.50) (3, 0.53) (4, 0.51) (5, 0.50)
				(6, 0.51) (7, 0.50) (8, 0.51) (9, 0.51) (10, 0.52)
			};
			\addlegendentry{Regular GAT (2 heads)}
			
			\addplot[color=orange,mark=diamond*] coordinates {
				(1, 0.30) (2, 0.31) (3, 0.31) (4, 0.30) (5, 0.31)
				(6, 0.30) (7, 0.31) (8, 0.29) (9, 0.29) (10, 0.29)
			};
			\addlegendentry{Regular GAT (1 head)}
			
		\end{axis}
    \end{tikzpicture}
    \caption{Time comparision for three different models compared to the NVGCN. As the plat shows, the NVGCN has spent the least time compared to other methods.}
    \label{tab:time_comp_NVGCN}
\end{table}

\section{Conclusion}
This paper demonstrates that the integration of Graph Neural Networks with Delaunay triangulations represents an advancement in the field of image classification. While this approach presents challenges, especially with more complex datasets, its computational efficiency and robustness benefits are shown on standard benchmarks. Combining GCNs with Delaunay triangulation graphs improves the accuracy and efficacy of image classification and opens new avenues for research in computer vision and image processing. This synergy holds significant potential to advance the field, encouraging further exploration and refinement of graph-based methods for image analysis, as discussed below.


Integrating GCNs with advanced techniques is promising but poses notable challenges, particularly in handling the computational complexity of large-scale graph representations. The quest to optimally blend these techniques for diverse image types and classification tasks also remains a fertile ground for research. Future efforts could be directed towards developing more efficient algorithms for graph construction and feature extraction, as well as investigating the adaptability of these methods across various image classification scenarios, including the real-time processing and classification of dynamic images.

Moreover, this research can be extended in several ways. One significant direction involves experimenting with various metrics or distance functions to construct Delaunay triangulation graphs. Another promising area of exploration is to demonstrate that employing Delaunay triangulation graphs can facilitate faster graph convolution steps when used in conjunction with GCNs. This could pave the way for more streamlined and efficient graph-based image processing techniques.\\

\noindent \textbf{Acknowledgments:}
This research work has been partially supported by the Natural Sciences and Engineering Research Council of Canada, NSERC, and the Faculty of Science, University of Windsor.

\bibliographystyle{IEEEtran}  

\bibliography{reference}
\newpage
\clearpage
\appendix 

\input{Mohammadi-appendix-NormalizedGCN_And_KCenter_Problem.tex}

\end{document}

%% file: Mohammadi-appendix-NormalizedGCN_And_KCenter_Problem.tex
\hyphenation{op-tical net-works semi-conduc-tor IEEE-Xplore}


\title{Appendix}

\date{Sep 29th, 2024}

\maketitle

\section*{The Normalized Graph Convolution Network}
\subsection*{Calculating the computational complexity of the normalized VGCN compared to conventional GCN}
In this section, we demonstrate how to calculate the time complexity in asymptotic notation in addition to the number of multiplications reduced by the proposed method.

\subsubsection{Asymtotic Notation Calculation}
In the main Graph Convolutional Network (GCN) formula, the degree matrix plays a crucial role, especially in the normalization step. Here's the time complexity analysis while explicitly incorporating the degree matrix.

A common version of the GCN layer formula is:

\[
H^{(l+1)} = \sigma \left( \hat{D}^{-\frac{1}{2}} \tilde{A} \hat{D}^{-\frac{1}{2}} H^{(l)} W^{(l)} \right)
\]

\noindent where:
\begin{itemize}
	\item \( \tilde{A} \) is the adjacency matrix with added self-loops, of size \( [M \times M] \).
	\item \( \hat{D} \) is the degree matrix, where \( \hat{D}_{ii} = \sum_j {A}_{ij} \), i.e., the degree of each node.
	\item \( H^{(l)} \) is the node feature matrix, of size \( [M \times F] \), where \( M \) is the number of nodes and \( F \) is the number of features.
	\item \( W^{(l)} \) is the trainable weight matrix for layer \( l \), of size \( [F \times F'] \), where \( F' \) is the number of output features.
	\item \( \sigma \) is a non-linear activation function, which doesn’t affect the time complexity.
\end{itemize}

We account for the time complexity of the three main matrix operations in this formula:

\begin{enumerate}
	\item \textbf{Degree matrix multiplication:} \( \hat{D}^{-\frac{1}{2}} H^{(l)} \)
	\item \textbf{Adjacency matrix multiplication:} \( \tilde{A} H^{(l)} \)
	\item \textbf{Weight matrix multiplication:} \( H^{(l)} W^{(l)} \)
\end{enumerate}

\textbf{ 1. Degree Matrix Multiplication:} \( \hat{D}^{-\frac{1}{2}} H^{(l)} \)

The degree matrix \( \hat{D} \) is a diagonal matrix. Multiplying a diagonal matrix with another matrix is relatively simple: each row of the feature matrix \( H^{(l)} \) is scaled by the corresponding diagonal element in \( \hat{D}^{-\frac{1}{2}} \).

- \( \hat{D}^{-\frac{1}{2}} \) is of size \( [M \times M] \); however,t since it is diagonal, this operation only takes \( O(N F) \) time, because there are \( M \) rows and each row involves \( F \) multiplications.

\textbf{2. Adjacency Matrix Multiplication:} \( \hat{A} H^{(l)} \)

The adjacency matrix \( \tilde{A} \) is of size \( [M \times M] \), and \( H^{(l)} \) is of size \( [M \times F] \). The matrix multiplication \( \tilde{A} H^{(l)} \) will take \( O(M^2 F) \) in a dense graph.

However, in a sparse graph with \( E \) edges, the complexity would be \( O(E F) \), because we only need to process the non-zero entries of the adjacency matrix (i.e., the edges). Figure \ref{fig:TMA_IMG} illustrates that the number of nodes in the Delaunay triangulation is much less than the number of pixels in an image. Also, in Figure \ref{fig:adj_TMA_img}, the adjacency matrix has been shown as evidence of the sparsity of the related graph.

\textbf{3. Second Degree Matrix Multiplication:} \( \hat{D}^{-\frac{1}{2}} (\tilde{A} H^{(l)}) \)

This is similar to the first degree matrix multiplication, and it takes \( O(M F) \), because it involves element-wise multiplication of each row of the resulting matrix by the corresponding degree.

\textbf{4. Weight Matrix Multiplication:} \( H^{(l)} W^{(l)} \)

After aggregating the neighborhood features, the matrix \( H^{(l)} \) is multiplied by the weight matrix \( W^{(l)} \). This is a standard matrix multiplication where:
- \( H^{(l)} \) is of size \( [M \times F] \).
- \( W^{(l)} \) is of size \( [F \times F'] \).

This operation takes \( O(M F F') \), because for each node, we perform \( F \times F' \) multiplications to transform the feature matrix.

\textbf{Total Time Complexity}

After the analysis performed above, we combine all these operations to compute the total time complexity of a GCN layer:

\begin{itemize}
	\item \textbf{Degree matrix multiplication:} \( O(M F) \)
	\item \textbf{Adjacency matrix multiplication:}
	\begin{itemize}
		\item \textbf{Dense case:} \( O(M^2 F) \)
		\item \textbf{Sparse case:} \( O(E F) \)
	\end{itemize}
	\item \textbf{Second degree matrix multiplication:} \( O(M F) \)
	\item \textbf{Weight matrix multiplication:} \( O(M F F') \)
\end{itemize}

Thus, the total time complexity for one GCN layer is as follows:

\begin{itemize}
	\item \textbf{Dense case:}
	
	\begin{align*}
		&O(M F) + O(M^2 F) + O(M F) + O(M F F') \\ 
		& = O(M^2 F + M F F')    
	\end{align*}

	\item \textbf{Sparse case:}
	\begin{align*}
		&O(M F) + O(E F) + O(M F) + O(M F F') \\
		& =O(E F + M F F')
	\end{align*}
\end{itemize}

\textbf{Final Time Complexity:}

\begin{itemize}
	\item \textbf{Dense graphs:} \( O(M^2 F + M F F') \)
	\item \textbf{Sparse graphs:} \( O(E F + M F F') \), where \( E \) is the number of edges.
\end{itemize}

These complexities account for all the matrix multiplications, including those involving the degree matrix. The time complexity would scale linearly for multiple layers with the number of layers \( L \). Thus, for \( L \) layers, the total complexity would be:

\begin{itemize}
	\item \textbf{Dense case:} \( O(L (M^2 F + M F F')) \)
	\item \textbf{Sparse case:} \( O(L (E F + M F F')) \)
\end{itemize}

If we were working with a \textbf{Delaunay Tessellation} graph, which is a type of sparse graph where the average degree of nodes is approximately 6.25, the time complexity for applying this graph structure in a Graph Neural Network (GNN) takes advantage of the sparsity. 

\textbf{Delaunay Tessellation Graph Properties}
A Delaunay tessellation graph has the following useful properties:
\begin{itemize}
	\item The \textbf{average degree} of each node is approximately 6.25, meaning that each node is connected to approximately 6 other nodes.
	\item The number of edges \( E \) in the graph is proportional to the number of nodes \( M \), i.e., \( E = O(M) \), because Delaunay tessellation graphs are sparse.
\end{itemize}

\textbf{Applying a Delaunay Tessellation Graph in GNNs}
The GNN formula for this case becomes:
\[
H_{D.T}^{(l+1)} = \sigma(\hat{A}_{D.T} H_{D.T}^{(l)} W^{(l)})
\]
\noindent where:
\begin{itemize}
	\item \( \hat{A}_{D.T} \) is the adjacency matrix with adding self-loop as $I\times 6$.
	\item \( H_{D.T}^{(l)} \) is the node feature matrix.
	\item \( W^{(l)} \) is the weight matrix.
	\item \( \sigma \) is the non-linear activation function.
\end{itemize}

\textbf{Step-by-Step Time Complexity}

\textbf{1. Adjacency Matrix Multiplication:} \( \hat{A}_{D.T} H^{(l)} \)
\begin{itemize}
	\item The adjacency matrix \( \hat{A}_{D.T} \) is sparse with \( E \approx O(M) \) edges, so this multiplication takes \( O(M F) \) time.
\end{itemize}

\textbf{2. Weight Matrix Multiplication:} \( H_{D.T}^{(l)} W^{(l)} \)
\begin{itemize}
	\item After adjacency matrix multiplication, we perform the feature transformation, multiplying \( H_{D.T}^{(l)} \) by \( W^{(l)} \), which takes \( O(M F F') \) time.
\end{itemize}

\textbf{Total Time Complexity (Without Degree Matrix Normalization)}
Without the degree matrix normalization, the total time complexity for one GNN layer simplifies to:
\[
O(M F) + O(M F F') = O(M F + M F F')
\]

\textbf{Time Complexity for Multiple Layers}
If we apply this simplified GNN across \( L \) layers, the total time complexity becomes:
\[
O(L (M F + M F F'))
\]

\textbf{Conclusion}
We significantly reduce the complexity by removing the degree matrix normalization for Delaunay tessellation graphs, which are inherently sparse and balanced in terms of node degrees. The resulting GNN time complexity is \( O(M F + M F F') \) per layer, or \( O(L (M F + M F F')) \) for \( L \) layers.

\subsubsection{Multiplication Calculation}
To compute the percentage reduction in the number of multiplications by removing the degree matrix normalization from the GNN, we analyze the number of multiplications in both cases: \textit{with} and \textit{without} the degree matrix normalization.

\textbf{1. Number of Multiplications With Degree Matrix Normalization}

The GNN formula with degree matrix normalization is:

\[
H^{(l+1)} = \sigma \left( \hat{D}^{-\frac{1}{2}} \tilde{A} \hat{D}^{-\frac{1}{2}} H^{(l)} W^{(l)} \right)
\]

This involves the following matrix multiplications:
\begin{itemize}
	\item \textbf{First Degree Matrix Multiplication} \( \hat{D}^{-\frac{1}{2}} H^{(l)} \): This requires \( O(M F) \) multiplications.
	\item \textbf{Adjacency Matrix Multiplication} \( \tilde{A} H^{(l)} \): For a sparse graph like Delaunay tessellation, this requires \( O(M F) \) multiplications.
	\item \textbf{Second Degree Matrix Multiplication} \( \hat{D}^{-\frac{1}{2}} (\tilde{A} H^{(l)}) \): This requires \( O(M F) \) multiplications.
	\item \textbf{Weight Matrix Multiplication} \( H^{(l)} W^{(l)} \): This requires \( O(M F F') \) multiplications, where \( F \) is the number of input features and \( F' \) is the number of output features.
\end{itemize}

Thus, the total number of multiplications with the degree matrix is:
\[
O(M F + M F F')
\]

\textbf{2. Number of Multiplications Without Degree Matrix Normalization}

When we remove the degree matrix normalization, the formula simplifies to:

\[
H_{D.T}^{(l+1)} = \sigma(\hat{A}_{D.T} H_{D.T}^{(l)} W^{(l)})
\]

This involves:
\begin{itemize}
	\item \textbf{Adjacency Matrix Multiplication} \( \hat{A}_{D.T} H_{D.T}^{(l)} \): For a sparse graph, this requires \( O(M F) \) multiplications.
	\item \textbf{Weight Matrix Multiplication} \( H_{D.T}^{(l)} W^{(l)} \): This requires \( O(M F F') \) multiplications.
\end{itemize}

Therefore, the total number of multiplications without the degree matrix is:
\[
O(M F + M F F')
\]

\textbf{3. Percentage Reduction in Multiplications}

The reduction in the number of multiplications is:

\begin{align*}
	&\text{Reduction}=\\
	& (3 M F + M F F') - (M F + M F F') =\\ 
	&2 M F
\end{align*}

The percentage reduction is:
\begin{align*}
	&\text{Percentage Reduction} =\\
	&\frac{2 M F}{3 M F + M F F'} \times 100
\end{align*}

This simplifies to:
\begin{align*}
	&\text{Percentage Reduction} =\\ &\frac{2}{3 + F'/F} \times 100
\end{align*}

\textbf{Interpretation}:

The percentage reduction depends on the ratio \( \frac{F'}{F} \), where \( F' \) is the number of output features and \( F \) is the number of input features:

- \textbf{When \( F' = F \)}: The output features are equal to the input features. In this case, the percentage reduction is:
\begin{align*}
	&\text{Percentage Reduction} =\\ 
	& \frac{2}{3 + 1} \times 100 = \\
	&\frac{2}{4} \times 100 = 50\%
\end{align*}
This means removing the degree matrix leads to a 50\% reduction in multiplications.

- \textbf{When \( F' < F \)}: The output features are fewer than the input features. As \( F' \) becomes smaller, the term \( \frac{F'}{F} \) decreases, and the percentage reduction increases. Some examples are:
\begin{itemize}
	\item When \( F' = \frac{F}{2} \) (half the input features):
	\begin{align*}
		&\text{Percentage Reduction} =\\ 
		&\frac{2}{3 + \frac{1}{2}} \times 100 = \frac{2}{3.5} \times 100 = 57.14\%
	\end{align*}
	\item When \( F' = \frac{F}{3} \) (one-third of the input features):
	\begin{align*}
		&\text{Percentage Reduction} =\\ 
		&\frac{2}{3 + \frac{1}{3}} \times 100 \approx 60\%
	\end{align*}
	\item As \( F' \ll F \), the percentage reduction approaches:
	\begin{align*}
		&\lim_{F' \to 0} \frac{2}{3 + \frac{F'}{F}} \times 100 =\\
		& \frac{2}{3} \times 100 = 66.67\%
	\end{align*}
\end{itemize}
In this scenario, removing the degree matrix becomes more impactful, leading to a higher percentage reduction in the number of multiplications.

- \textbf{When \( F' > F \)}: The output features are greater than the input features. As \( F' \) becomes larger, the term \( \frac{F'}{F} \) increases, causing the percentage reduction to decrease. Some examples are:
\begin{itemize}
	\item When \( F' = 2F \) (twice the input features):
	\[
	\text{Percentage Reduction} = \frac{2}{3 + 2} \times 100 = \frac{2}{5} \times 100 = 40\%
	\]
	\item When \( F' = 3F \) (three times the input features):
	\[
	\text{Percentage Reduction} = \frac{2}{3 + 3} \times 100 = \frac{2}{6} \times 100 = 33.33\%
	\]
	\item As \( F' \gg F \), the percentage reduction approaches:
	\[
	\lim_{F' \to \infty} \frac{2}{3 + \frac{F'}{F}} \times 100 = 0\%
	\]
\end{itemize}

As the number of output features \( F' \) grows larger than the number of input features \( F \), the degree matrix multiplications become less significant compared to the weight matrix multiplication, leading to a smaller reduction in the total number of multiplications.

\begin{figure}
	\centering
	\includegraphics[width=1\linewidth]{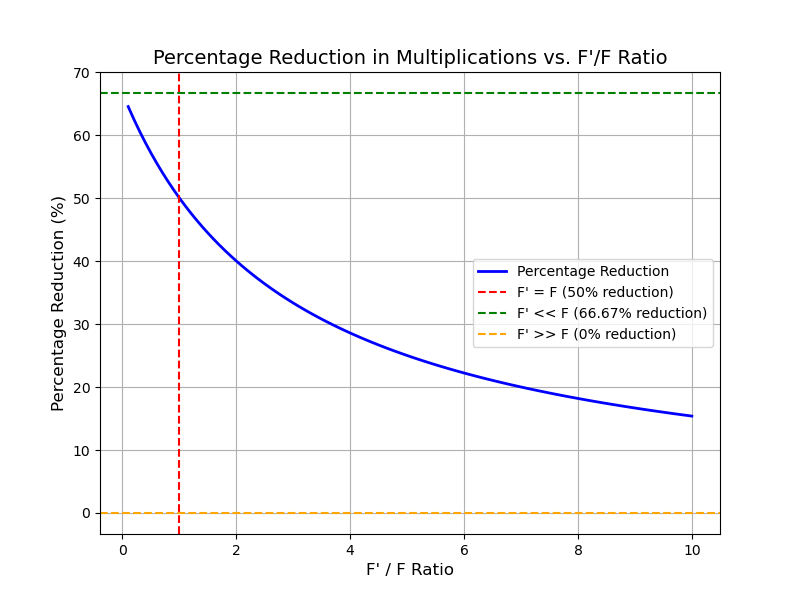}
	\caption{Percentage Reduction in Multiplications vs. F'/F Ratio }
	\label{fig:Multiplication_Ratio}
\end{figure}

The exact reduction depends on the ratio of \( F' \) to \( F \), but in most practical cases, it will lie between \( 33\% \) and \( 66.67\% \). Figure \ref{fig:Multiplication_Ratio} shows the percentage reduction in multiplications compared to different F'/F ratio.

\newpage

	\section*{Optimal Number of Regions in Image Segmentation is NP-Hard}
	
	Determining the optimal number of regions in image segmentation is a fundamental challenge in computer vision and image processing. The goal is to partition an image into regions (clusters) such that pixels within the same region are similar in some sense (e.g., color, texture, spatial proximity), and pixels in different regions are dissimilar. However, finding the best number of regions that minimizes intra-region variance and maximizes inter-region variance is computationally intensive. This complexity arises because the number of possible ways to partition an image increases exponentially with the number of pixels, making exhaustive search infeasible for practical purposes.
	
	This problem is inherently NP-Hard due to its combinatorial nature. Selecting the optimal segmentation involves exploring an exponential search space to find the global minimum of an objective function, which is computationally prohibitive. As a result, approximate algorithms and heuristics are typically employed to find near-optimal solutions within reasonable computational time.
	
	\subsection*{Reduction to the K-Center Problem}
	
	The difficulty of finding the optimal number of regions in image segmentation can be formally demonstrated by reducing it to the well-known \textit{K-Center Problem}, which is NP-Hard. The K-Center Problem aims to select \( k \) center points to minimize the maximum distance between any data point and its nearest center.
	
	\textbf{Definition:}
	
	Given a set of points \( P = \{p_1, p_2, \dots, p_n\} \) in a metric space with distance function \( d(\cdot, \cdot) \), the objective is to find a set of centers \( C = \{c_1, c_2, \dots, c_k\} \subseteq P \) that minimizes the following objective function:
	
	\[
	\min_{C} \max_{p \in P} \min_{c \in C} d(p, c)
	\]
	
	\subsection*{Mapping Image Segmentation to the K-Center Problem}
	
	In image segmentation, each pixel \( p_i \) can be represented in a feature space that captures both spatial and color information. For instance, a pixel can be represented as:
	
	\[
	p_i = (x_i, y_i, l_i, a_i, b_i)
	\]
	
	where \( (x_i, y_i) \) are the spatial coordinates, and \( (l_i, a_i, b_i) \) are the color components in the Lab color space. The distance between two pixels can be defined as:
	
	\begin{align*}
	&d(p_i, p_j) = \sqrt{DISTs}
    \\
    &which
    \\
    &DISTs = \lambda_s \left( (x_i - x_j)^2 + (y_i - y_j)^2 \right) + \\
    &\lambda_c \left( (l_i - l_j)^2 + (a_i - a_j)^2 + (b_i - b_j)^2 \right)
    \end{align*}

	where \( \lambda_s \) and \( \lambda_c \) are weighting factors that balance the importance of spatial and color differences.
	
	\subsection*{Objective Function in Image Segmentation}
	
	The segmentation problem aims to assign each pixel \( p_i \) to a region represented by a center \( c_j \) such that the maximum distance from any pixel to its assigned center is minimized:
	
	\[
	\min_{C} \max_{p_i \in P} \min_{c_j \in C} d(p_i, c_j)
	\]
	
	This objective mirrors that of the K-Center Problem, where we seek centers that minimize the maximum intra-cluster distance.
	
	\subsection*{NP-Hardness Proof through Reduction}
	By mapping the image segmentation problem to the K-Center Problem, we establish that:
	
	\begin{enumerate}
		\item \textbf{Instance Equivalence:} An instance of the image segmentation problem can be transformed into an instance of the K-Center Problem by representing pixels as points in a metric space with the defined distance function.
		\item \textbf{Solution Correspondence:} A solution to the K-Center Problem provides a set of centers that correspond to the centroids of image regions in the segmentation problem.
		\item \textbf{NP-Hardness Transfer:} Since the K-Center Problem is NP-Hard, and a polynomial-time solution to the image segmentation problem would imply a polynomial-time solution to the K-Center Problem, it follows that the image segmentation problem is also NP-Hard.
	\end{enumerate}
	
	\subsection*{Conclusion}
	
	Therefore, finding the optimal number of regions in image segmentation is NP-Hard because it can be reduced to the K-Center Problem. This complexity justifies the use of approximate algorithms like SNIC (Simple Non-Iterative Clustering) superpixel generation, which provides efficient and practical solutions without guaranteeing global optimality.


%% file: main.bbl
\begin{thebibliography}{10}
\providecommand{\url}[1]{#1}
\csname url@samestyle\endcsname
\providecommand{\newblock}{\relax}
\providecommand{\bibinfo}[2]{#2}
\providecommand{\BIBentrySTDinterwordspacing}{\spaceskip=0pt\relax}
\providecommand{\BIBentryALTinterwordstretchfactor}{4}
\providecommand{\BIBentryALTinterwordspacing}{\spaceskip=\fontdimen2\font plus
\BIBentryALTinterwordstretchfactor\fontdimen3\font minus \fontdimen4\font\relax}
\providecommand{\BIBforeignlanguage}[2]{{%
\expandafter\ifx\csname l@#1\endcsname\relax
\typeout{** WARNING: IEEEtran.bst: No hyphenation pattern has been}%
\typeout{** loaded for the language `#1'. Using the pattern for}%
\typeout{** the default language instead.}%
\else
\language=\csname l@#1\endcsname
\fi
#2}}
\providecommand{\BIBdecl}{\relax}
\BIBdecl

\bibitem{zhou2020graph}
J.~Zhou, T.~Cui, and et~al., ``Graph neural networks: A review of methods and applications,'' \emph{AI Open}, vol.~1, pp. 57--81, 2020.

\bibitem{mallat1989theory}
S.~Mallat and et~al., ``A theory for multiresolution signal decomposition: the wavelet representation,'' \emph{IEEE Transactions on Pattern Analysis and Machine Intelligence}, vol.~11, no.~7, pp. 674--693, 1989.

\bibitem{8-SLIC-Achanta2012SLIC}
``Slic superpixels compared to state-of-the-art superpixel methods,'' \emph{IEEE Transactions on Pattern Analysis and Machine Intelligence}, vol.~34, no.~11, pp. 2274--2282, 2012.

\bibitem{15-SNIC-achanta2017superpixels}
R.~Achanta and et~al., ``Superpixels and polygons using simple non-iterative clustering,'' in \emph{Proceedings of the IEEE Conference on Computer Vision and Pattern Recognition}, 2017, pp. 4651--4660.

\bibitem{9-turbopixel}
A.~Levinshtein and et~al., ``Turbopixels: Fast superpixels using geometric flows,'' \emph{IEEE Transactions on Pattern Analysis and Machine Intelligence}, vol.~31, no.~12, pp. 2290--2297, 2009.

\bibitem{11-LinearSpectralClustering7814265}
J.~Chen and et~al., ``Linear spectral clustering superpixel,'' \emph{IEEE Transactions on Image Processing}, vol.~26, no.~7, pp. 3317--3330, 2017.

\bibitem{6-kumar2023extensive}
B.~Kumar and Vinoth., ``An extensive survey on superpixel segmentation: A research perspective,'' \emph{Archives of Computational Methods in Engineering}, pp. 1--19, 2023.

\bibitem{4-aurenhammer1991voronoi}
F.~Aurenhammer, ``Voronoi diagrams — a survey of a fundamental geometric data structure,'' \emph{ACM Computing Surveys (CSUR)}, vol.~23, no.~3, 1991.

\bibitem{12-digitized_vor_tess}
C.~Thanh-Tung, E.~Herbert, and et~al., ``Triangulations from topologically correct digital voronoi diagrams,'' \emph{Computational Geometry}, vol.~48, no.~7, pp. 507--519, 2015.

\bibitem{kcenterProblem}
J.~Garcia-Diaz and et~al., ``Approximation algorithms for the vertex k-center problem: Survey and experimental evaluation,'' \emph{IEEE Access}, vol.~7, pp. 109\,228--109\,245, 2019.

\bibitem{13-kmeans-lloyd1982least}
S.~Lloyd, ``Least squares quantization in pcm,'' \emph{IEEE Transactions on Information Theory}, vol.~28, no.~2, pp. 129--137, 1982.

\bibitem{14-kmeans-macqueen1967some}
M.~James. and et~al., ``Some methods for classification and analysis of multivariate observations,'' in \emph{Proceedings of the fifth Berkeley Symposium on Mathematical Statistics and Probability}, vol.~1, no.~14, 1967, pp. 281--297.

\bibitem{HierarchicalClustering_Mittal2022}
H.~Mittal and et~al., ``A comprehensive survey of image segmentation: clustering methods, performance parameters, and benchmark datasets,'' \emph{Multimedia Tools and Applications}, vol.~81, no.~24, pp. 35\,001--35\,026, 2022.

\bibitem{FuzzyClustering_YANG19931}
S.~Yang, ``A survey of fuzzy clustering,'' \emph{Mathematical and Computer Modelling}, vol.~18, no.~11, pp. 1--16, 1993.

\bibitem{GNNModelScarselli2009TheGN}
S.~Franco, G.~Marco, and et~al., ``The graph neural network model,'' \emph{IEEE Transactions on Neural Networks}, vol.~20, pp. 61--80, 2009.

\bibitem{18-MessagePassing-Gilmer2017NeuralMP}
G.~Justin, S.~Samuel, and et~al., ``Neural message passing for quantum chemistry,'' in \emph{ICML}, 2017.

\bibitem{19-GAT-Velickovic2018GraphAN}
V.~Petar, C.~Guillem, and et~al., ``Graph attention networks,'' in \emph{ICLR}, 2018.

\bibitem{16-DouglasPeuker}
P.~Douglas and H.~David, ``Algorithms for the reduction of the number of points required to represent a digitalized line or its caricature,'' \emph{Cartographica: The International Journal for Geographic Information and Geovisualization}, vol.~10, no.~2, pp. 112--122, 1973.

\bibitem{kipf2017semi}
T.~N. Kipf and M.~Welling, ``Semi-supervised classification with graph convolutional networks,'' \emph{arXiv preprint arXiv:1609.02907}, 2017.

\bibitem{book_deBerg2008}
M.~de~Berg and et~al., \emph{Computational Geometry: Algorithms and Applications}, 3rd~ed.\hskip 1em plus 0.5em minus 0.4em\relax Springer, 2008.

\bibitem{20-pygeometric-Fey2019PyTorch}
M.~Fey and et~al., ``Fast graph representation learning with {PyTorch Geometric},'' 2019.

\bibitem{Wavelet_vasudevan2023image}
V.~Vasudevan and et~al., ``Image classification using graph neural network and multiscale wavelet superpixels,'' \emph{Pattern Recognition Letters}, vol. 166, pp. 89--96, 2023.

\bibitem{RAG_avelar2020superpixel}
P.~Avelar and et~al., ``Superpixel image classification with graph attention networks,'' in \emph{2020 33rd SIBGRAPI Conference on Graphics, Patterns and Images (SIBGRAPI)}, 2020, pp. 203--209.

\end{thebibliography}
